\documentclass[num-refs]{wiley-article}
\usepackage{color}
\usepackage[normalem]{ulem}  
\usepackage{amsfonts} % added for mathbb

\usepackage{multirow}
\usepackage[scr=esstix,cal=boondox]{mathalfa} 
\usepackage[ruled,vlined]{algorithm2e}

\renewcommand{\v}[1]{\ensuremath{\mathbf{#1}}}
\newcommand{\gv}[1]{\bm{#1}}

\usepackage[nice]{nicefrac}

\papertype{Research Article}

\begin{document}
\runningauthor{Sacco M. A. et al. }
\title{Evaluation of Machine Learning Techniques for Forecast Uncertainty Quantification}

\author[1,2]{Maximiliano A. Sacco}
\author[2,4,5]{Juan J. Ruiz}
\author[5,6]{Manuel Pulido}
\author[3]{Pierre Tandeo}
%\author[3]{Yicun Zhen}

\affil[1]{Servicio Meteorol\'ogico Nacional, Buenos Aires, Argentina}
\affil[2]{Universidad de Buenos Aires, Facultad de Ciencias Exactas y Naturales, Departamento de Ciencias de la Atm\'osfera y los Oc\'eanos. Buenos Aires, Argentina}
\affil[3]{Lab-STICC, UMR CNRS 6285, IMT Atlantique, Plouzan\'e, France}
\affil[4]{CONICET – Universidad de Buenos Aires. Centro de Investigaciones del Mar y la Atm\'osfera (CIMA), Buenos Aires, Argentina}
\affil[5]{CNRS – IRD – CONICET – UBA. Instituto Franco-Argentino para el Estudio del Clima y sus Impactos (IRL 3351 IFAECI). Buenos Aires, Argentina}
\affil[6]{Departamento de F\'isica - Facultad Ciencias Exactas y Naturales y Agrimensura, Universidad Nacional del Nordeste, Corrientes, Argentina}

\corraddress{msacco@smn.gob.ar}
\corremail{msacco@smn.gob.ar}
\maketitle
%\brieftitle{Machine Learning for Uncertainty Quantification}
%\runninghead
\begin{abstract}
Ensemble forecasting  is, so far, the most successful approach to produce relevant forecasts with an estimation of their uncertainty. The main limitations of ensemble forecasting are the high computational cost and the difficulty to capture and quantify different sources of uncertainty, particularly those associated with model errors. 
In this work we perform toy-model and state-of-the-art model experiments to analyze to what extent artificial neural networks (ANNs) are able to model the different sources of uncertainty present in a forecast. In particular those associated with the accuracy of the initial conditions and those introduced by the model error. We also compare different training strategies: one based on a direct training using the mean and spread of an ensemble forecast as target, the other ones rely on an indirect training strategy using an analyzed state as target in which the uncertainty is implicitly learned from the data. Experiments using the Lorenz'96 model show that the ANNs are able to emulate some of the properties of ensemble forecasts like the filtering of the most unpredictable modes and a state-dependent quantification of the forecast uncertainty. Moreover, ANNs provide a reliable estimation of the forecast uncertainty in the presence of model error. 
Preliminary experiments conducted with a state-of-the-art forecasting system also confirm the ability of ANNs to produce a reliable quantification of the forecast uncertainty.

\keywords{neural networks, chaotic dynamic models, uncertainty quantification, observation likelihood loss function, forecast}

\end{abstract}

\section{Introduction}
\label{SEC:INTRO}

Uncertainty in weather forecasting is attributed to errors in the initial conditions and model errors. The first ones are associated with imperfect knowledge of the initial state of the system. The last ones are produced by truncation errors and an inaccurate representation of small-scale processes in numerical forecast models. While model errors can be reduced by a higher horizontal resolution and better physical parameterizations (i.e., improving the model), initial condition errors can be reduced by improving observation networks and data assimilation systems \citep{carrassi2018}.

Currently, the uncertainty associated with weather forecasts is usually quantified by ensemble forecasts. To produce an ensemble forecast, several numerical simulations are performed using different initial conditions appropriately selected \citep[e.g.][]{kalnay2003} and using different approaches to represent the imperfection in the model such as parametric error representation and stochastic processes  \citep[e.g.][]{Hagedorn2012}. One of the main limitations of ensemble forecasting systems lies in its high computational cost and in the complexity of designing them to adequately capture and quantify the different sources of uncertainty, particularly model errors. 

As stated by \cite{vannitsemetal2021} systematic errors in the forecasted values as well as deficiencies in its uncertainty quantification can be alleviated by different forecast post-processing approaches. These techniques aim to provide better guidance for objective decision-making based on forecast information.

According to \cite{vannitsemetal2021} there are an increasing number of works using machine learning as a post-processing tool with promising results. Several techniques, such as analog regression \citep{hamill2006,lguensat2018}, random forests \citep{taillardat2016,schlosser2019} and \citep{herman2018}, neural networks \citep{Rasp2018,bremmes2020}, fuzzy neural networks \citep{lu2021} and variational autoencoders \citep{grooms2021,minahandgrooms2021}, have been successfully applied for the increase of the forecast accuracy as well as to provide a more reliable quantification of the forecast uncertainty. 

Artificial neural networks (ANNs) are used to reduce systematic errors, particularly in situations in which the systematic error relationship with the state of the system is highly nonlinear \citep[e.g.][]{hauptetal2021,marzban2003}. \cite{farchietal2021,bonavitaetal2020} use machine learning to represent model errors. In particular, \cite{farchietal2021} use data assimilation to update the machine learning model parameters as new observations became available. 

Recently, neural network representations have been proposed to jointly estimate the systematic component of the forecast error and the forecast uncertainty. \cite{scher} designed a methodology  based on a convolutional neural network that learned the standard deviation of the error from an existing ensemble system. The convolutional neural networks predict the spatial distribution of the forecast error variance using a deterministic forecast as input. The learning target is the error variance estimated from an ensemble forecast. The skill of the network on forecasting the uncertainty is lower than the one obtained with an ensemble of forecasts, but the network provides a reasonable quantification of the uncertainty at a much lower computational cost than those of ensemble forecasts. Furthermore, the network performs better than other low-computational cost approaches, such as a simple estimation of the uncertainty based on a multiple linear regression. Along the same lines, \cite{gronquist2021} used a neural network to improve the uncertainty quantification obtained with a limited size ensemble. In this case, a deep neural network is used to model the ensemble spread generated with a large ensemble using a small-sized ensemble as input. They show a significant improvement in the skill of probabilistic forecasts and particularly for extreme weather events. One disadvantage of these approaches is that the network models the uncertainty estimated by an ensemble of forecasts, which can be biased, particularly in the presence of model errors. 

A different approach estimates the forecast uncertainty directly from the  data without requiring an a-priori estimation of the forecast uncertainty such as that provided by an ensemble of forecasts. \cite{D_Isanto_2018,camporeale2,Rasp2018,veldkampetal2021,barnesandbarnes2020} and \cite{wang} introduced neural network architectures and training methodologies that directly quantify the uncertainty performing the training over a large database of forecasts and their corresponding observations. 

The key aspect is the definition of a loss function which should have a well-defined sensitivity to the state-dependent forecast uncertainty. \cite{wang,barnesandbarnes2020} achieved this by defining a loss function inspired in the observation likelihood, while \cite{camporeale1}, \cite{D_Isanto_2018}, \cite{Rasp2018} and \cite{veldkampetal2021} define a loss function based on the continuous ranked probability score (CRPS) and the Brier score. These works conclude that this is a feasible approach for a cost-effective estimation of the forecast uncertainty for idealized and real-data cases. In these approaches the shape of the probability distribution of the forecast error has to be assumed (usually a Gaussian distribution), so,  they can be considered as parametric or distribution-based approaches.

\cite{scheuerer2020,clare2021} propose a non-parametric approach to estimate forecast uncertainty purely from data and without making assumptions on the shape of the underlying forecast error distribution. They discretize forecasted variables into bins and predict the probability of the forecasted value to fall within each bin, providing a discrete approximation of the probability density function. Along similar lines, \cite{cannon2018,gasthaus19a,bremnes2020} use neural networks and quantile-regression to quantify the forecast uncertainty without assuming the shape of the forecast error distribution. 

Along the line of these works which have shown how the use of machine learning techniques and neural networks can successfully perform uncertainty estimation from a deterministic forecast, the objective of this work is to analyze to what extent a neural network is able to identify the different sources of uncertainty present in a forecast, in particular those associated with the initial conditions and those introduced by the model error. To achieve this goal, experiments were conducted in a perfect model scenario, where the source of uncertainty is only introduced by the initial conditions, and an imperfect model scenario where model error is also a source of uncertainty. These experiments were repeated for different forecast lead times.
We use three different parametric approaches to estimate the uncertainty using neural networks whose performances are investigated and compared. Proof-of-concept realistic experiments are conducted to evaluate the performance of the methodologies using data from a state-of-the-art numerical weather prediction system.

This work is organized as follows: Section \ref{SEC:UQ} presents an introduction to the problem of uncertainty quantification in the forecasting of chaotic systems, and the notation used in this work. The architecture of the neural networks and the proposed training methodology is described in section \ref{SEC:METH}. A detailed description of the design of all the synthetic experiments involved in this work is presented in section \ref{SEC:EXP}. Section \ref{SEC:RES} analyzes the obtained results. A simple experiment with real data was carried out in section \ref{SEC:RDC} and the conclusions of this work are finally presented in section \ref{SEC:CONC}.

\section{Forecast uncertainty quantification}
\label{SEC:UQ}
The evolution of a dynamical system, like the atmosphere or the ocean, is represented in this work via a Markov process
\begin{equation}
\label{EQU:sdm1}
\v x_{k} = \mathcal{M}_{k:k-1}(\v x_{k-1})+\gv \eta_k,
\end{equation}
where $\v x_{k}$ is an $S$-dimensional state vector of  the  system at time $k$, $\mathcal{M}_{k:k-1}$ is a known surrogate model of the system dynamics (e.g., a numerical weather prediction model or a climate simulation model) that propagates the state of the system from time $k-1$ to time $k$.  This model is assumed to be nonlinear with chaotic properties. The random model errors, $\gv\eta_k$, are samples from an unknown probability distribution. Since we consider systematic model errors, this distribution is expected to have a non-zero mean.  
The evolution of the system results from the recursive application of the model over several time steps. In this case, model errors interact with the system dynamics:

\begin{equation}
\label{EQU:sdm2}
\v x_{k+l} = \underbrace{\mathcal{M}(\mathcal{M}(\cdots (\mathcal{M}}_{l-\text{times}}(\v x_{k}) +  \gv\eta_{k+1})+\gv\eta_{k+2})+...)+\gv\eta_{k+l}.
\end{equation}
% which can be simplified as:
In this work, these interactions are simplified to a representation with an additive state-dependent error term 
\begin{equation}
\label{EQU:sdm3}
\v x_{k+l} =  \mathcal{M}_{k+l:k}(\v x_{k}) + \gv\delta_{k+l:k}(\v x_{k},\gv\eta_{k+1},...,\gv\eta_{k+l}),
\end{equation}
where $\gv\delta_{k+l:k}$ represents the accumulated effect of model errors and its evolution under the nonlinear system dynamics between times $k$ and $k+l$ and $\mathcal{M}_{k+l:k}$ is the nonlinear operator resulting from the $l$-times recursive application of the surrogate model. 
For numerical weather prediction, a deterministic forecast ($\v x^d$) is usually obtained by integrating the dynamical model starting from the most probable initial condition usually referred as the analysis ($\v x^a$), such that

\begin{equation}
\label{EQU:detfor}
\v x^{d}_{l,k+l} =  %\mathcal{M}_{k+l:k}(\v x^{d}_{0,k}),
\mathcal{M}_{k+l:k}(\v x^{a}_{k}),
\end{equation}
where $\v x^{d}_{l,k+l}$ is the forecasted state at lead time $l$ (first subindex) valid at time $k+l$ (second subindex) and $\v x^{a}_{k}$ is the analysis at time $k$ (i.e., a pointwise estimation of the system state at time $k$ given all the available observations up to time $k$). This estimation is often obtained using data assimilation \citep[e.g.][]{kalnay2003,carrassi2018}. 

The resulting forecast error is defined as
\begin{equation}
\label{EQU:forerr}
\gv \epsilon_{l,k+l}^{d} = \v x^{t}_{k+l} - \v x^{d}_{l,k+l},
\end{equation}
where $\v x^{t}_{k+l}$ is the actual state of the system usually referred to as the true state. 
The forecast error arises due to the presence of errors in the initial conditions and in the model formulation. Initial condition errors are driven by the dynamical system. Furthermore, model error is also partially driven by the dynamical system as explicitly noted in Eq. \ref{EQU:sdm2}. This effect is state-dependent, meaning that error growth rate depends on the state of the system which is a particularly challenging aspect of short and medium range prediction of atmospheric phenomena. The main goal of this work is to  model the state-dependent mean ($\overline{\epsilon}_{l,k+l}$) and standard deviation ($\sigma_{l,k+l}$) of the forecast error distribution using machine learning techniques. Hereinafter, we drop the time index (second subscript) whenever possible and only retain the forecast lead time as subindex to simplify the notation (i.e., $\v x^{d}_{l}=\v x^{d}_{l,k+l}$).
 
Currently, the most common way to estimate the state dependent mean and covariance of the forecast error distribution is through ensemble forecasting. In this approach, the initial state is assumed uncertain and, therefore, given by a probability distribution. This initial distribution is represented through a sample, i.e., an ensemble of initial states. Then, an ensemble of model integrations is performed initialized from the different initial conditions. In the case of  nonlinear dynamical models, these integrations spread from each other. The magnitude of the spread is used to estimate the forecast standard deviation at a given lead time. The $n-$th ensemble member ($x^{e,(n)}$) is generated by integrating the model from the $n-$th  initial condition ,$x^{a,(n)}$, 
\begin{align}
  \label{EQU:MODEL1}
  \v x^{e,(n)}_{l} =  \mathcal{M}^{(n)}_{l:0}(\v x^{a,(n)}) + \hat{\gv\delta}^{(n)}_{l:0},
\end{align}
where $\v x^{e,(n)}_{l}$ is the forecasted state vector given by the $n-th$ ensemble member at lead time $l$, $\hat{\gv\delta}^{(n)}_{l:0}$ is  a representation of the effect of model errors, and $\mathcal{M}^{(n)}_{l:0}$ is the numerical model. Note that a different numerical model can be used for the integration of each ensemble member as in stochastically perturbed parameters or parametrization outputs.  
Given an ensemble of forecast states (i.e., a sample of possible future states of the system), a pointwise estimate of the state of the system at lead time $l$ can be obtained from the ensemble mean state vector
\begin{align}
  \label{EQU:MODEL2}
  \overline{\v x}^{e}_{l} = \frac{1}{N} \sum_{n=1}^{N} \v x^{e,(n)}_{l},
\end{align}
where $N$ is the number of ensemble members.
This estimate is statistically more accurate in terms of the root mean square error, than the one provided by a deterministic forecast \citep{kalnay2003}. The forecast uncertainty is usually quantified in high-dimensional systems through the second moment by computing the sample error covariance matrix,
\begin{equation}
\label{EQU:SIGMA}
\gv \Sigma^{e}_{l} = \frac{1}{(N-1)}\sum_{n=1}^{N} \left(\v x^{e,(n)}_{l}-\overline{\v x}^{e}_{l}\right)\left(\v x^{e,(n)}_{l}-\overline{\v x}^{e}_{l}\right)^\top .
\end{equation}
While the full covariance matrix is commonly used in data assimilation applications (see \cite{carrassi2018} for more details), most forecast applications are only concerned with the error associated with the forecasted variables so that there is no need to quantify  the correlation between the state variables and only the diagonal elements of the covariance matrix, ${\gv \sigma^{e}_{l}}^{2}=\mathrm{diag}(\Sigma^{e}_{l})$ are of interest. The accuracy of the sample covariance estimate depends on the ensemble size and on the adequacy of model error representation.

\section{Methodology}
\label{SEC:METH}

\begin{figure}[hbt!]
        \includegraphics[width=\textwidth]{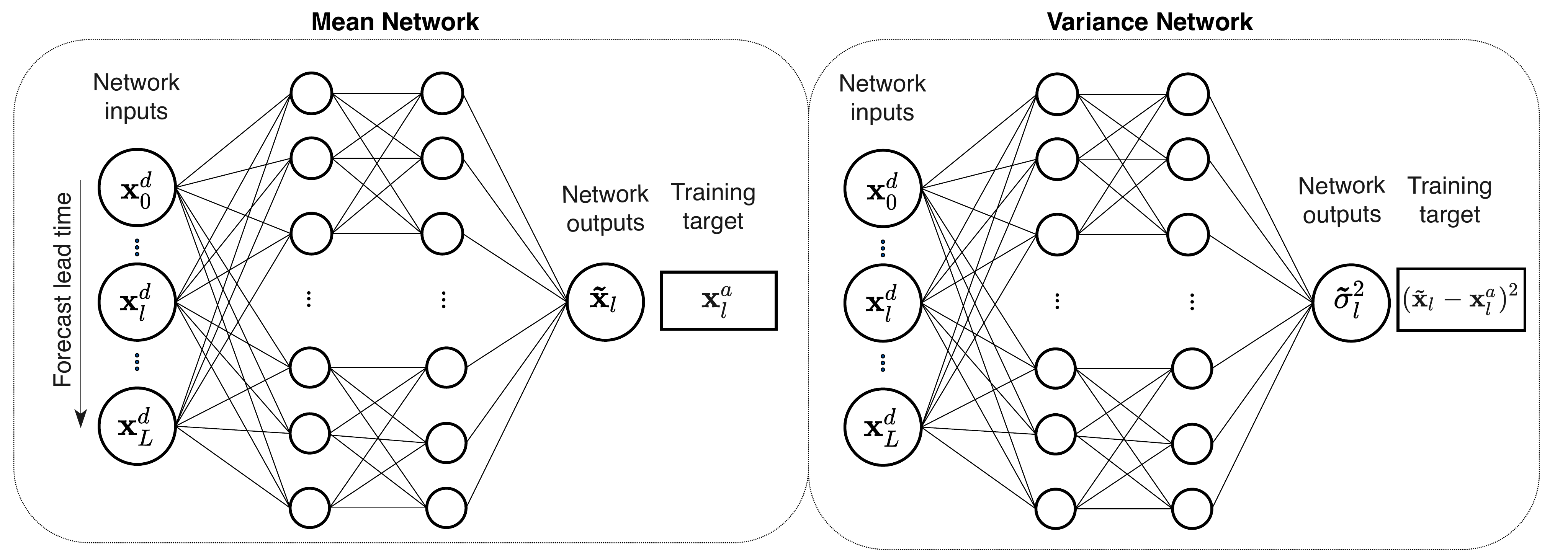}
        \caption{Schematic of the architecture of the network estimating the mean state of the system (left) and the network estimating the forecast error variance (right). First the network that provides the mean state is trained using the analysis as target. Then, the network estimating the forecast error variance is trained using a proxy of the local error variance such as the difference between the mean forecasted state and the analysis as target.}
\label{FIG:NET}
\end{figure}

Two ANNs are considered to provide the first and the second moments of the probability distribution of the predicted state of the system, which is assumed Gaussian. As input, both networks (Fig. \ref{FIG:NET}) use a deterministic forecast of the state of the system at different lead times ($\{\v x^{d}_{0},\ldots,\v x^{d}_{l},\ldots,\v x^{d}_{L}\}, 0 \leq l \leq L$ ), providing information about the dynamic evolution of the system over a time window surrounding lead time $l$. Our goal is to provide the predicted mean state of the system ${\tilde{\v x}_{l}}$ (implicitly the systematic component of forecast error, since a forecast with the physical model is used as input), and to estimate the state-dependent variance of the forecast error, ${\tilde{\gv\sigma}}^2$. Note that both ${\tilde{\v x}_{l}}$ and ${\tilde{\gv\sigma}}^2$ are S-dimensional vectors. 

Predictions of ${\tilde{\v x}_{l}}$ are required to estimate the heterosdastic noise of the forecast ${\tilde{\gv\sigma}}^2_l(\v x_{l})$, and at the same time the heterosdastic noise affects the ${\tilde{\v x}_{l}}$ regression. Thus, in general, a two-way coupling between the two quantities is present so that both should be estimated jointly. In this work,  we assume homoscedastic noise for the mean state regression, in other words, we assume ${\tilde{\gv\sigma}}^2$  does not affect ${\tilde{\v x_{l}}}$ estimation. Under this assumption, the two neural networks which represent the functional dependences of ${\tilde{\v x}_{l}}$ and ${\tilde{\gv\sigma}}^2$ as a function of the input, may be trained separately. We note that the homoscedastic noise assumption is only for the mean state regression, for the variance regression the noise is assumed heterosdastic.

The mean state neural network is trained first, and once convergence is reached, the variance neural network, which depends on the mean state predictions, is trained. This simplification contributes to a more reliable convergence of the neural networks and also to the interpretation of the results. Furthermore, it enables the use of several alternative loss functions. However, we note that the variance state dependence may genuinely affect the mean state estimation as shown in the seminal work \citep{nix94}. The importance of capturing or neglecting this effect may depend on the context.

\subsection{Training the mean state network}

First, the mean state network is trained as shown in Figure \ref{FIG:NET}.
We use the S-dimensional analysis states $(\v x^a_l)$ valid at the same time of the forecast $(\tilde{\v x}_l)$ as the training target and a loss function based on the mean square error (MSE)

\begin{equation}
\label{EQU:MSEM}
\mathcal{L}^{x}_{MSE}(\tilde{\v x}_l;\v x^a_l)=(\tilde{\v x}_l - \v x^a_l)^\top(\tilde{\v x}_l - \v x^a_l).
\end{equation}

The analysis states are used as targets, rather than the available observations, because they provide a complete estimation of the state vector on the model grid at regular times and with an error which is statistically smaller than the observational error \citep{carrassi2018}. This scheme directly provides an estimation of the mean state in a supervised manner.

\subsection{Training the variance network}
\subsubsection{Training with the ensemble spread (\textit{NN-mse})} 
\label{SEC:MSE}

In this strategy, we follow a similar approach to \cite{scher} where ${\tilde{\gv \sigma}_{l}}^2$ is estimated with a network trained using as target a dataset of ${{\gv \sigma}_{l}^e}^2$  (i.e. the sample forecast variance determined by an ensemble of forecast at lead time $l$). As for the mean state network, the ${\tilde{\gv \sigma}_{l}}^2$ and ${\gv \sigma_l^e}^2$ are assumed to be S-dimensional. In this scheme the variance network learns the dependence of variance with the mean forecast in a fully  supervised manner.

Network training is performed using a MSE-based loss function which is defined as follows:
\begin{equation}
\label{EQU:MSES}
\mathcal{L}^{\sigma}_{MSE} ({\tilde{\gv \sigma}_l}^2;{\gv \sigma_l^e}^2)= ({\tilde{\gv \sigma}_l}^2 - {\gv \sigma_l^e}^2)^\top({\tilde{\gv \sigma}_l}^2 - {\gv \sigma_l^e}^2).
\end{equation}

The main drawback of using  supervised training is that it requires an \textit{explicit uncertainty} data set (i.e.,  ${\sigma^{e}_{l}}^2$) which in turn needs a full ensemble forecast integration to provide an estimation of the forecast error variance for each element in the training sample. Furthermore, it will inherit the limitations of the ensemble forecast.

\subsubsection{Training with a locally estimated error (\textit{NN-ext})} 
\label{SEC:eMSE}
A second approach assumes the only information available for training are the analysed states ($\v x^{a}_{l}$). the only information used for training are the analysed states ($\v x^{a}_{l}$). We refer to this kind of approach as \textit{indirect training}, because no a-priori information of the variance of the forecast error is required.  This is estimated directly from the data \citep[e.g.][]{camporeale1,wang}. Indirect training approaches require a loss function definition that is sensitive to the variance of the forecast error (i.e. an uncertainty aware loss function). To construct such a loss function,  first a local proxy of the corrected forecast error is obtained,

\begin{equation}
\label{EQU:eMSE}
\tilde {\gv \epsilon}_{l} =  \tilde{\v x}_l-\v x^a_l.
\end{equation}
where $\tilde{\v x}_l$ is the output of the mean state network.
Next, the element-wise squared error (i.e. $\tilde {\gv \epsilon}_{l}^2 $ ) is used as a local estimate of the variance of the corrected forecast. In this case, the loss function is defined as,
\begin{equation}
\label{EQU:eMSE2}
\begin{aligned}
\mathcal{L}^{\sigma}_{eMSE}(\tilde{\gv \sigma}_l;\tilde {\gv \epsilon}_{l})&= (\tilde{\gv \sigma}_l^2 -  \tilde {\gv \epsilon}_{l}^2   )^\top(\tilde{\gv \sigma}_l^2 -  \tilde {\gv \epsilon}_{l}^2 ),
\end{aligned}
\end{equation}
where the squares denote element-wise or Hadamard products, i.e. $\tilde {\gv \epsilon}_{l}^2\doteq \tilde {\gv \epsilon}_{l} \odot \tilde {\gv \epsilon}_{l}$. We named this loss function \textit{extended MSE (eMSE)}, since we use the MSE metric in combination with a target based on a local estimation of variance 

\subsubsection{Training based on the sample likelihood (\textit{NN-lik})} 
\label{SEC:NLE}
Another form of indirect training can be obtained assuming that the error in the corrected forecast ($\tilde{\v x}$) follows a Gaussian distribution with zero mean and covariance $\tilde{\Sigma}$. We approximate the covariance matrix as a diagonal matrix,
\begin{equation}
\label{EQU:NLESigma}
\tilde \Sigma_l  \approx  \mathrm{diag}\left( \tilde{\gv \sigma}_l^2 \right), 
\end{equation}

The diagonal assumption significantly simplifies the definition of $\tilde{\gv\Sigma}$ and the computation. This is taken because  only the quantification of the error variance is usually required for forecasting applications.  

The loss function based on the negative of the log-likelihood is given by
\begin{equation}
\label{EQU:NLPME}
\mathcal{L}^{\sigma}_{lik}(\tilde{\v \sigma}_l;\tilde {\gv \epsilon}_{l} ) =  \log \left( \det \left( \tilde {\gv\Sigma}_l \right) \right)  +  \tilde {\gv \epsilon}_{l}^\top  \tilde{\gv \Sigma}_l^{-1} \tilde {\gv \epsilon}_{l} 
\end{equation}
The variance $\tilde{\gv \sigma}^2$ can be estimated as the one that minimizes the loss function, and so it maximizes the log-likelihood of the analyzed states over the entire dataset \citep{wang,Pulido2018,Tandeo2020}.

By introducing this loss function, we are assuming independence of the errors at different model variables and times for this forecasting application. Moreover, the analysis uncertainty is not explicitly taken into account. As with the NN-ext (sec. \ref{SEC:eMSE}), this loss function has the advantage of using an \textit{implicit uncertainty} dataset and then there is no need for an a priori estimation of the forecast uncertainty.

In this section, we defined different loss functions considering a single sample of the dataset. However, it should be kept in mind that the optimization of the ANNs aims to minimize the loss function of the entire dataset, which is given by the sum over all the samples in the training dataset.

Equations \ref{EQU:eMSE2} and \ref{EQU:NLPME} have different shapes as a function of the variance, and so different conditionings of the optimization problem result that might affect the convergence rate. Therefore, we decided to evaluate the performance in the training of both loss functions.

\section{Experimental design}

\label{SEC:EXP}
Experiments using a simple chaotic dynamical system are conducted to evaluate the performance of the ANNs. Two different scenarios are considered, a scenario in which the model used to estimate the state of the system and to produce forecasts is a perfect representation of the true system dynamics (i.e., $\eta_k = 0$ in Eq.\ref{EQU:sdm1}) and another scenario in which the model is a surrogate model in which small-scale dynamics are not represented and, therefore, forecasts are affected by model errors.

\subsection{Perfect model scenario}
\label{sec:PMS}
In the perfect-model scenario (PMS)  we use the single-scale version of Lorenz'96 equations \citep{lorenz96} given by
\begin{equation}
\frac{dx_{(i)}}{dt}=-x_{(i-2)}x_{(i-1)}+x_{(i-1)}x_{(i+1)}-x_{(i)}+F,
\label{Lorenz1S}
\end{equation}
where $x_{(i)}$ is the $i-th$ component of $\v x$ and $F$ is a constant that represents an external forcing.  
Cyclic boundary conditions are imposed so that this set of equations resembles an advective-dissipative forced system over a circle of latitude.

Firstly, a long integration is performed with the model, which is assumed to be the unknown true evolution of the system. The model is integrated  during 500 time units using the 4th order Runge-Kutta method and a forcing of $F=8$, a time-step of 0.0125 time units and $S=8$ state variables. This simulation will be referred to as the \textit{nature} integration. Secondly, a set of noisy observations are generated from the \textit{nature} integration by adding an uncorrelated Gaussian noise with zero mean and covariance $\v R=\sigma_R^2 \v I$ with $\sigma_R=1$. All the state variables are observed every 4 time-steps (0.05 time units).

Initial conditions for the numerical forecasts are generated using an ensemble-based data assimilation system \citep[the Local Ensemble Transform Kalman Filter, LETKF; ][]{HUNT07}, which provides an analysis every time observations are available (i.e., every 0.05 time units). An ensemble of 50 members is used to generate the analysis. The data assimilation is conducted over the length of the \textit{nature} integration giving a total of 13000 estimations of the state of the system and its associated uncertainty (7000 used as training set, 3000 for validation and 3000 for testing). 
Finally, two sets of forecasts are generated, an ensemble forecast in which each member is initialized from the members of the analysis ensemble (to be used in the direct training approach) and a deterministic forecast initialized from the mean of the analysis ensemble (to be used as input to the neural network). The target used for the training of the ANNs is the mean of the ensemble analysis states. 
The forecasts are integrated up to a maximum lead time of 3.5 time units (280 time steps). This maximum lead time is taken because the forecast error is close to its nonlinear saturation at this lead time.
 In all the network training synthetic experiments, true trajectories are assumed to be unknown and forecasts and analyses from the data assimilation system are used as inputs and targets respectively, considering what would be available in realistic experiments.
\subsection{Imperfect model scenario}

For the generation of the \textit{nature} integration in the imperfect model scenario (IMS), we use the two-scale Lorenz' model \citep{lorenz96} which is defined by the following system of coupled differential equations:
\begin{equation}
\begin{split}
  \frac{dx_{(i)}}{dt}= &-x_{(i-1)}(x_{(i-2)}-x_{(i-1)})-x_{(i)}+F-\frac{hc}{b}\sum_{j=J(i-1)+1}^{iJ} y_{(j)}
  \label{Lorenz2Sa}
  \end{split}
\end{equation}
\begin{equation}
\begin{split}
  \frac{dy_{(j)}}{dt}= & -cb \,y_{(j+1)}(y_{(j+2)}-y_{(j-1)})-c \,y_{(j)}+\frac{hc}{b}x_{(\mathrm{int}[(j-1)/J]+1)},
  \end{split}
\label{Lorenz2Sb}
\end{equation}
where $x_{(i)}$ are the state variables associated to the slow dynamics, while $y_{(j)}$ are the variables associated with the faster dynamics. $J$ is the number of $y$ variables for each $x$ variable, and $h$, $c$ , and $b$ are time-independent parameters controlling the coupling strength between the two systems. Cyclic boundary conditions are applied to both sets of equations, namely $x_{(1)}=x_{(S+1)}$, and $y_{(1)}=y_{(J\cdot S+1)}$. This two-scale system is a simple representation of systems with multiple spatio-temporal scales like the atmosphere or the ocean. In our experiments, J=32 and S=8 (i.e., the $y$ vector has a total of 256 variables) and the forcing term $F$ is set to 20 to obtain a chaotic behavior. 

The surrogate model used for data assimilation and forecasting is given by
\begin{equation}
\frac{dx_{(i)}}{dt}=-x_{(i-1)}(x_{(i-2)}-x_{(i-1)})-x_{(i)}+F+ G_{(i)}, 
\label{EQU:Lorenz2SPar1}
\end{equation}
where $G_{(i)}$ is a state dependent parametrization term that approximates the effect of the missing dynamics (i.e., the effect of fast variables $y$). As in \cite{pulido16}, $G_{(i)}$ is assumed to be a linear function of the state variable $x_{(i)}$ 
\begin{equation}
G_{(i)}=\alpha x_{(i)} + \beta , 
\label{EQU:Lorenz2SPar2}
\end{equation}
with $\alpha=19.16$ and $\beta=-0.81$ constant parameters whose optimal values are taken from \cite{scheffler2019}.
The use of a surrogate model which does not consider the small-scale dynamics allows to consider the effects of model error in the forecasts.
The general experimental setting including the observation frequency, observation noise, and length of the assimilation experiment are the same as in the PMS. The ensemble forecasts generated in the imperfect model scenario do not include an explicit representation of model errors (i.e., $\hat{\delta}^{(n)}_{l:0} = 0$ in  Eq.\ref{EQU:sdm3}), all ensemble members are integrated with the same model configuration version.

\subsection{Network architecture and training}

A fully connected architecture was chosen for all the experiments in this work because it is a generic architecture, meaning that it does not make any assumptions about the data, and because the low dimensionality of the problems we are analyzing allows it. For higher-dimensional problems, other architectures (e.g., convolutional) may be used to diminish the number of free network parameters.
For each experiment, both networks (mean state and variance) have two hidden layers with 50 neurons each and softplus activation functions \citep{softplus}.  
The mean state network should generate outputs within the range of possible state values, therefore, we use a linear activation function in the output layer. For the variance network, we use a softplus activation function in the output layer as the variances should be positive.  

The network size was defined based on preliminary experiments varying the size of the hidden layers (from 16 to 200 neurons) and the activation functions of the hidden layers (sigmoid, ReLU and softplus). The chosen architecture was the one with the lowest average loss function value over all the experiments.
All the networks were trained with the Adam optimizer \citep{adam} and using minibatches with a batchsize of $50$ elements. The loss function is evaluated over the validation set every 20 training epochs. The training stops when the loss function evaluated over the validation set stops decreasing or starts to increase (early stop). 
Early in the experimentation we noticed a great sensitivity of the training to the hyperparameters learning rate (LR) and weight decay (WD). So we conducted a brute force exploration of the parameter space using  LR=$\{1e^{-2},1e^{-3},1e^{-4},1e^{-5},1e^{-6}\}$ and   WD=$\{0,5e^{-1},1e^{-1},1e^{-2}\}$. The training of the NNs was repeated 5 times for each combination of LR and WD. The chosen hyperparameters are those that produce the lowest value of the cost function evaluated over the validation set, averaged over the 5 repetitions of the training.

\begin{figure}%[hbt!]
    \centering
        \includegraphics[width=.5\textwidth]{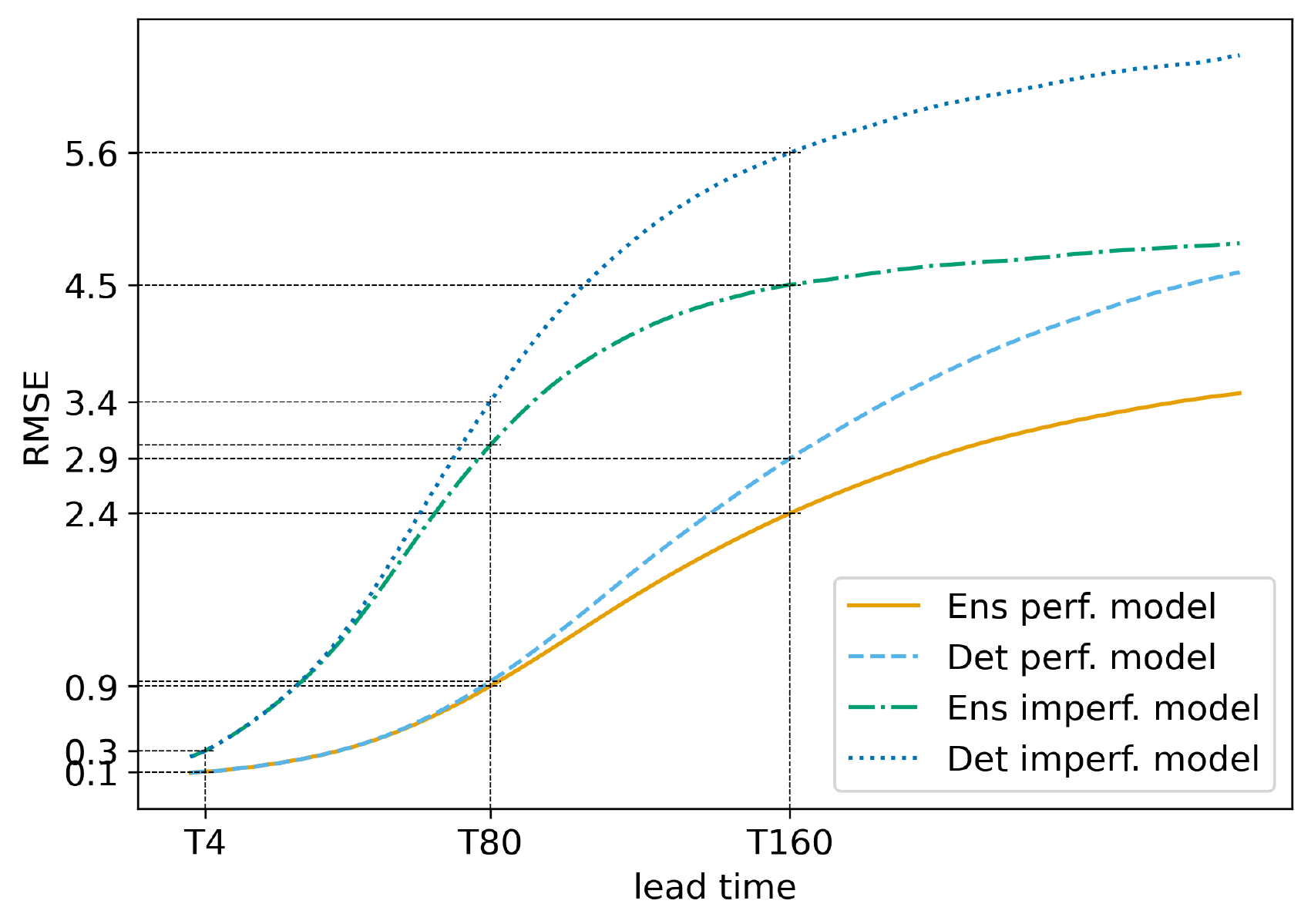}

        \caption{RMSE as a function of lead time in the perfect and imperfect model scenarios using the evolution of a single model integration (Det) and the mean of an ensemble of 50 integrations of the model (Ens). }
         \label{fig:RMSE_LEAD}

\end{figure}

For the evaluation of the performance of the  ANNs in the experiments, three different lead times were taken as representative, 4, 80 and 160 time steps, on the two  scenarios, PMS and IMS.
Figure \ref{fig:RMSE_LEAD} shows the evolution of deterministic and ensemble mean forecast errors for the PMS and IMS as a function of lead time. The comparison of the performance in different lead times is expected to shed some light in the impact of initial condition and model error uncertainties on the predicted neural network  forecast uncertainty. 

\label{SEC:ARCH}

The first lead time (T4) is of interest because it roughly corresponds to a very short range forecast (4 time steps), like those routinely used in the data assimilation cycle. At this lead time, the forecast error is comparable to the error present in the analyses used as training target by the ANN.  The second lead time (T80) is within the linear error growth regime, particularly in the perfect model experiment (see Fig. \ref{fig:RMSE_LEAD}). At this lead time, the difference between the error in the deterministic forecast and the ensemble forecast is small in the PMS but is important in the IMS. Unlike T4, the initial error propagated through the dynamics of the system at T80 has grown enough to become independent of the error present in the analyses. The third lead time (T160) is within the nonlinear error growth regime and it is also close to the error saturation in the IMS. In this lead time, the initial error has evolved through the dynamics of the system and has grown large enough to dominate the state of the system. Therefore, the system is at the limit of predictability regardless of the model error.

For each scenario (PMS and IMS) and for each selected lead time (T4, T80, T160), we conducted a deterministic forecast, an ensemble forecast (baseline for comparison) and three ANN predictions which were trained using the three loss functions described in Section \ref{SEC:METH} (i.e. NN-mse, NN-ext and NN-lik).

\subsection{Verification metrics}
\label{SEC:MET}
To compare different aspects of the performance of the different forecasting system (deterministic, ensemble, and ANNs with different training strategies) we use four metrics, root-mean-square error, coverage probability, standard deviation-error correlation and the flatness of the probability integral transform histogram. They are used as diagnostics in which the corresponding \textit{nature} state, $\v x^t$, is used as ground truth. %Evaluations using CPRS  showed that  the relative performance of the different networks as quantified by the CRPS score is similar to the one obtained by the RMSE (suggesting that in this case errors in the mean state dominates differences in the CRPS among the different forecasts).

\subsubsection{Root mean square error}
\label{rmse}
The purpose of this metric is to evaluate the accuracy in estimating the state of the system. The closer the value is to zero, the more accurate the forecast estimate.
The RMSE is computed as
\begin{equation}
\label{EQU:RMSE2}
\mathrm{RMSE} = \sqrt{ \frac{1}{S\cdot M} \sum_{l=1}^{M}(\tilde{\v x}_l - \v x^t_l)^\top(\tilde{\v x}_l - \v x^t_l) },
\end{equation}
where $\v x^t_l$ is the true state of the system corresponding to the forecast time, $S$ is the state vector dimension and $M$ is the total number of samples in the verification set. The RMSE is a deterministic-oriented metric \cite{murphy} and does not take into account the forecast uncertainty, so that  the main purpose of this metric in this work is to measure the accuracy of the corrected forecast ($\tilde{\v x}$). 

\subsubsection{Coverage Probability}
To evaluate the quality of the estimation of the standard deviation made by the network, we use the coverage probability. Given a confidence interval for the estimator, it consists of calculating the number of times that the confidence interval contains the true value,
\begin{equation}
\label{EQU:CP}
\mathrm{CP} = \frac{\sum_{l=0}^{M}\mathbb{1}(\mathcal{l}_l < x^a_l < \mathcal{u}_l)}{M},
\end{equation}
\label{CP}
where $\mathcal{l}_l$/$\mathcal{u}_l$ is the lower/upper bound of the confidence interval for an estimator with normal distribution $\mathcal{N}(\tilde{x}_l,\tilde{\sigma}_l)$, $\mathbb{1}$ is the indicator function (takes the value of 1 if the condition is true and 0 otherwise) and $M$ the size of the dataset. The perfect value for CP is equal to the confidence level used to construct the confidence interval, in our case 90\% or (0.9). A CP value higher/lower than the confidence level would indicate an over/underestimation of the uncertainty. In this work we use the coverage probability as a global measure of the reliability of the quantified uncertainty.

\subsubsection{Standard deviation-absolute error correlation}
\label{corr}

Since we aim to estimate the state dependent standard deviation of the forecast error, the Pearson's correlation coefficient is computed between the estimated forecast standard deviation $\tilde\sigma_{l}$ and the absolute value of the actual forecast error $ \left| \epsilon_l \right|= ( \left| \tilde{\v x}_l - \v x^t_l \right| $),
\begin{equation}
\mathrm{CORR}(\tilde {\sigma}_{l},  \left| \tilde\epsilon_{l}  \right| )=\frac{\mathrm{cov}(\tilde\sigma_{l},  \left| \epsilon_{l} \right| )}{\mathrm{std}(\tilde \sigma_{l})\mathrm{std}( \left| \epsilon_{l} \right| )}
\end{equation}
where $\mathrm{cov}$ and $\mathrm{std}$ are the spatio-temporal covariance and standard deviation respectively. 
The purpose of this metric is to evaluate the correspondence in the temporal evolution of the uncertainty and the error produced in the forecast. A good estimate of uncertainty should be correlated with the forecast error. Negative correlation values or values very close to zero could indicate poor estimation. 

\subsubsection{PIT Histograms}
Another important characteristic to analyze is the calibration of probabilistic forecasts. According to \cite{gneiting07} the calibration reflects the consistency between the estimated error probability distribution and the actual error probability distribution in an statistical sense. Calibration can be quantified using the probability integral transform histograms (PIT) as in \cite{gneiting07}. If $F$ is the predicted cumulative density function (CDF) of the forecast error ($\epsilon_{(i)}$) for a given state-variable $x_{(i)}$ at a particular time, then the PIT histogram is constructed as the histogram of $F_t(\epsilon^t_{(i)})$ (i.e. the CDF evaluated at the actual error, or at an appropriate proxy of it) over different times and state-variables. A flat histogram is an indication of consistency between the estimated and the actual error probability distributions. However, as discussed by \cite{hamill2001}, a flat PIT histogram is not sufficient to confirm a calibrated forecast, since a flat PIT histogram can result from the compensation of conditional biases. According to \cite{wilks2019} the flatness of the PIT histogram can be measured using the following metric:

\begin{equation}
{\chi}^2=\frac{N_b}{M}\sum_{b=1}^{N_b}(f_b-\frac{N_b}{M})^2,
\label{chisquare}
\end{equation}

where $f_b$ is the frequency at the $b^{th}$ bin, $N_b$ is the number of bins and $M$ is the sample size.  $\chi^2$ equal to 0 corresponds to a flat histogram while increasing positive values indicate increasing departure from the flat histogram.
\section{Results}
\begin{figure}[hbt!]
\centering
         \includegraphics[width=.5\textwidth]{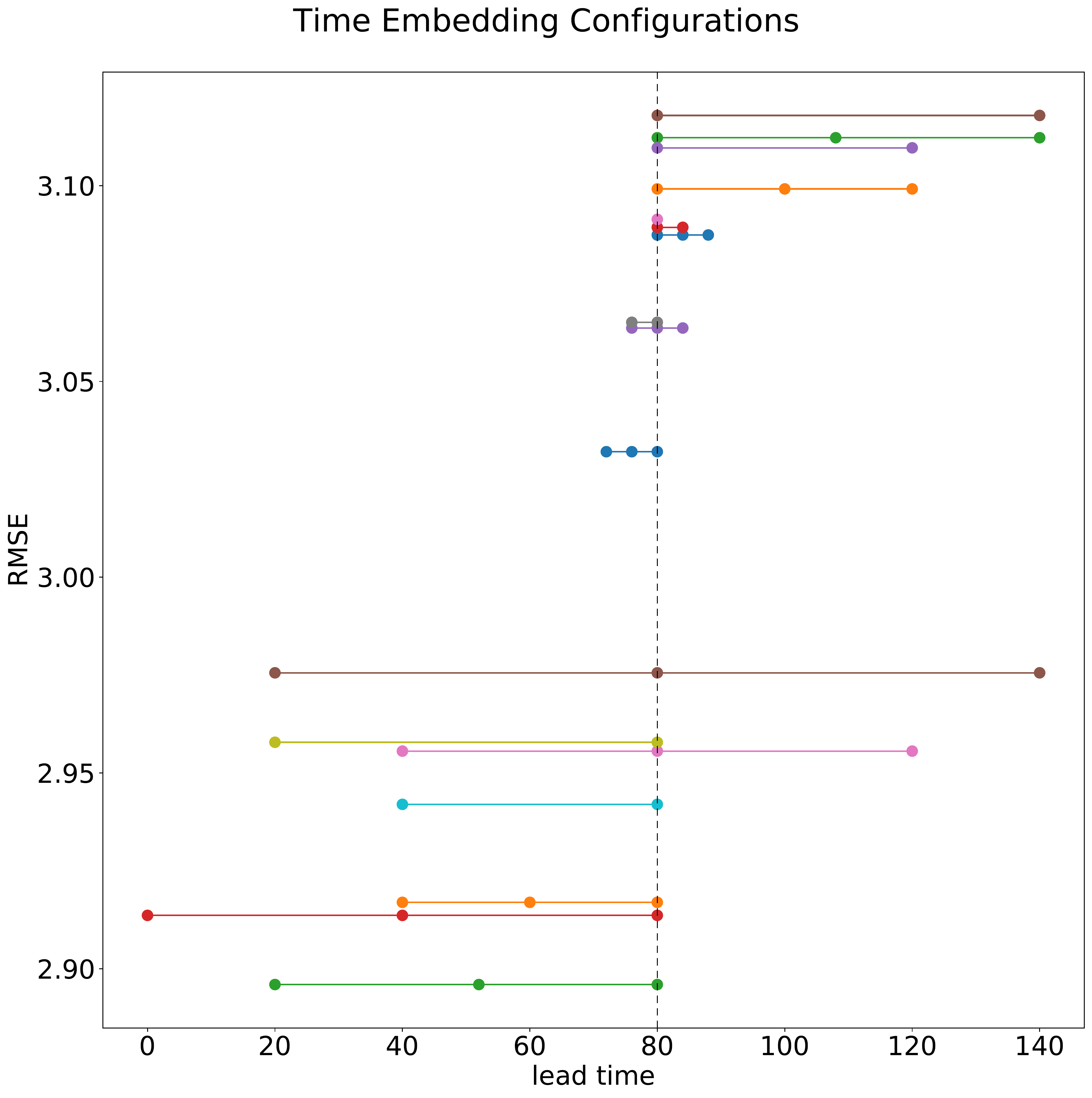}
    \caption{RMSE of the corrected forecast for experiments with different inputs. The RMSE is indicated in the $y$ axis, while the $x$ axis indicates the lead time. Each experiment is represented by a horizontal line with a different color, indicating the lead time range spanned by the input. The dots indicate the specific lead times aggregated to construct the input to the ANN. For example, the dark green line at the bottom of the figure has dots at lead times 20, 50 and 80 meaning that the input to the network resulted from the aggregation of the deterministic forecast at these three lead times. In this case, the RMSE achieved over the testing sample corresponding to this network is 2.9.}
    \label{fig:NetInSel}
\end{figure}
\label{SEC:RES}
\subsection{Sensitivity to network input}
\label{SEC:NIS}
We conducted experiments to investigate the sensitivity of the network performance to the forecast lead times included in the input. 
The input to the network consists of an aggregation of $L$ model states at different forecast lead times, with $L=1,2$ or $3$; i.e. $\{\v x_{l_1}\}$, $\{\v x_{l_1}, \v x_{l_2}\}$ or $\{\v x_{l_1}, \v x_{l_2}, \v x_{l_3}\}$, in which $l_j$ is representing the forecast lead time. 

We performed a set of 17 experiments combining different number of model states (from 1 to 3 lead times), lead time window lengths (from 4 to 60 time steps) and lead time shifts with respect to the output lead time, to evaluate how these different inputs influence the accuracy of the corrected forecast ($\tilde{\v x}$). 
In these experiments, the output variable is the model state at T80  lead time (i.e., $l=80$).
All the experiments were conducted in the imperfect model scenario using the same hyper-parameter and network architecture. For each experiment, we repeat the ANN training ten times using different initial weights to reduce the impact of weight initialization upon the resulting performance of the networks. 

Figure \ref{fig:NetInSel} shows the RMSE of the predicted state $\tilde{ \gv x}_l$ 
obtained with different inputs. The RMSE is indicated in the $y$ axis, while the $x$ axis indicates the lead time. Each experiment is represented by an horizontal line indicating the lead time range spanned by the input. The dots indicate the states at specific lead times aggregated to construct the input to the ANN.

In general, all the inputs performed reasonably well. However, the ones including information from lead times prior to the output lead time ($l$) perform better than the ones relying on lead times after $l$. This may be related with the fact that forecast errors, and particularly model errors, evolve according to the system dynamics (see Eq. \ref{EQU:sdm2}), thus including information about previous states can contribute to better describe the state dependent forecast mean error at lead time $l$. The range of lead times spanned by the input also produces a significant impact upon the ANN performance. Networks spanning a wider range of lead times (and using information from lead times prior to $l$) perform better. Furthermore, the performance of the ANN with the number of input variables  $L=3$ improves over the ones using $L=2$ and $L=1$. Best performance in the experiments is obtained for the ANN with three input variables at lead times T20, T50, and T80, i.e. $\{\v x^{d}_{20},\v x^{d}_{50},\v x^{d}_{80}\}$.

Based on these results, in the rest of this work, all the ANNs use the following inputs: for T4 the input consists of $\{\v x^{d}_{0},\v x^{d}_{4}\}$ (only two input lead times because not intermediate data would be available in realistic systems), for T80 we use $\{\v x^{d}_{0},\v x^{d}_{40},\v x^{d}_{80}\}$; and for T160, we use $\{\v x^{d}_{0},\v x^{d}_{80},\v x^{d}_{160}\}$.

\subsection{PMS experiments}
\label{SEC:RPMS}
\begin{figure}[hbt!]
    \centering
%    \begin{tabular}{cc}
     \includegraphics[width=\textwidth]{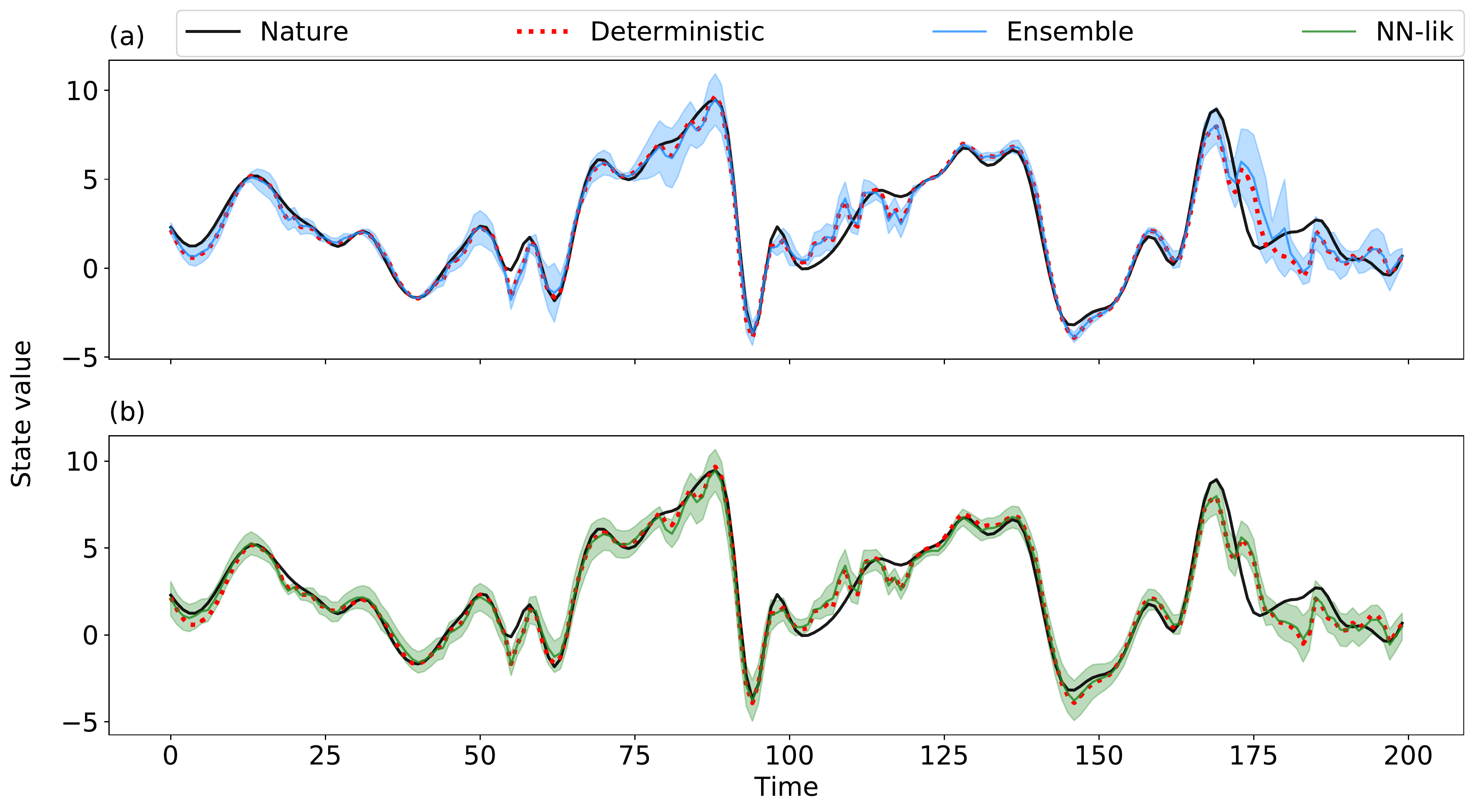}
%    \end{tabular}
    \caption{Time evolution of the first state variable in the T80 forecast experiment for 200 consecutive time steps taken from the testing set. (a) Mean of the ensemble forecasts (solid blue line) and the predicted spread (shaded). (b) Prediction with the NN-lik approach (solid green line) and the predicted spread (shaded). In both panels, the deterministic model forecast (red dotted lines) and the ground truth (solid black lines) are shown.}
    \label{fig:timelineSE}
    
\end{figure}

  In the perfect model experiments, forecast errors are the result of errors in the specification of the initial conditions. This particular scenario allows us to evaluate how well the ANNs capture error growth due to the chaotic behavior. Furthermore, these experiments are designed to evaluate whether ANNs are able to filter the most uncertain modes present in the forecast state as  in the case of the forecast ensemble mean  \citep[e.g.][]{kalnay2003}.  In ensemble forecasting, the ensemble mean $ \bar{\v x}^{e}_l$ exerts a filtering of the forecast error associated to predictability loss as $l$ increases. Moreover, the ensemble spread ($\gv \sigma^{e}_l$)  in the PMS is a reliable estimate of the state-dependent standard deviation of the forecast error. This is particularly true when the initial ensemble perturbations are generated using an ensemble Kalman filter \citep{candille}. 

Figure \ref{fig:timelineSE}a shows the time evolution of the ensemble mean forecast at T80 for the first state variable 
Figure \ref{fig:timelineSE}b shows the neural network prediction obtained in the NN-lik approach. 
At first glance, both the ensemble and the networks provide good estimations of the state of the system. Since we are in the linear error regime, in general, no big differences are found between the performance of the deterministic forecast and those from the ensemble mean and the network.
In terms of forecast uncertainty, overall, both systems provide a good estimation of the state dependent forecast uncertainty (i.e., the black curve is within the shaded area most of the time). The uncertainty estimated from the ANN  shows a smaller time variability than the one from the ensemble.  
The relationship between the forecast and its uncertainty is not trivial. Similar values of the deterministic forecast for the first Lorenz variable ($x^{d}_{(1)}$) are associated with different uncertainties. However, the periods of rapid increase or decrease in $x_{(1)}$ with time appears to be associated with lower forecast uncertainty in both, the ensemble and the ANN.

\begin{figure}[hbt!]
%    \centering
     \includegraphics[width=\textwidth]{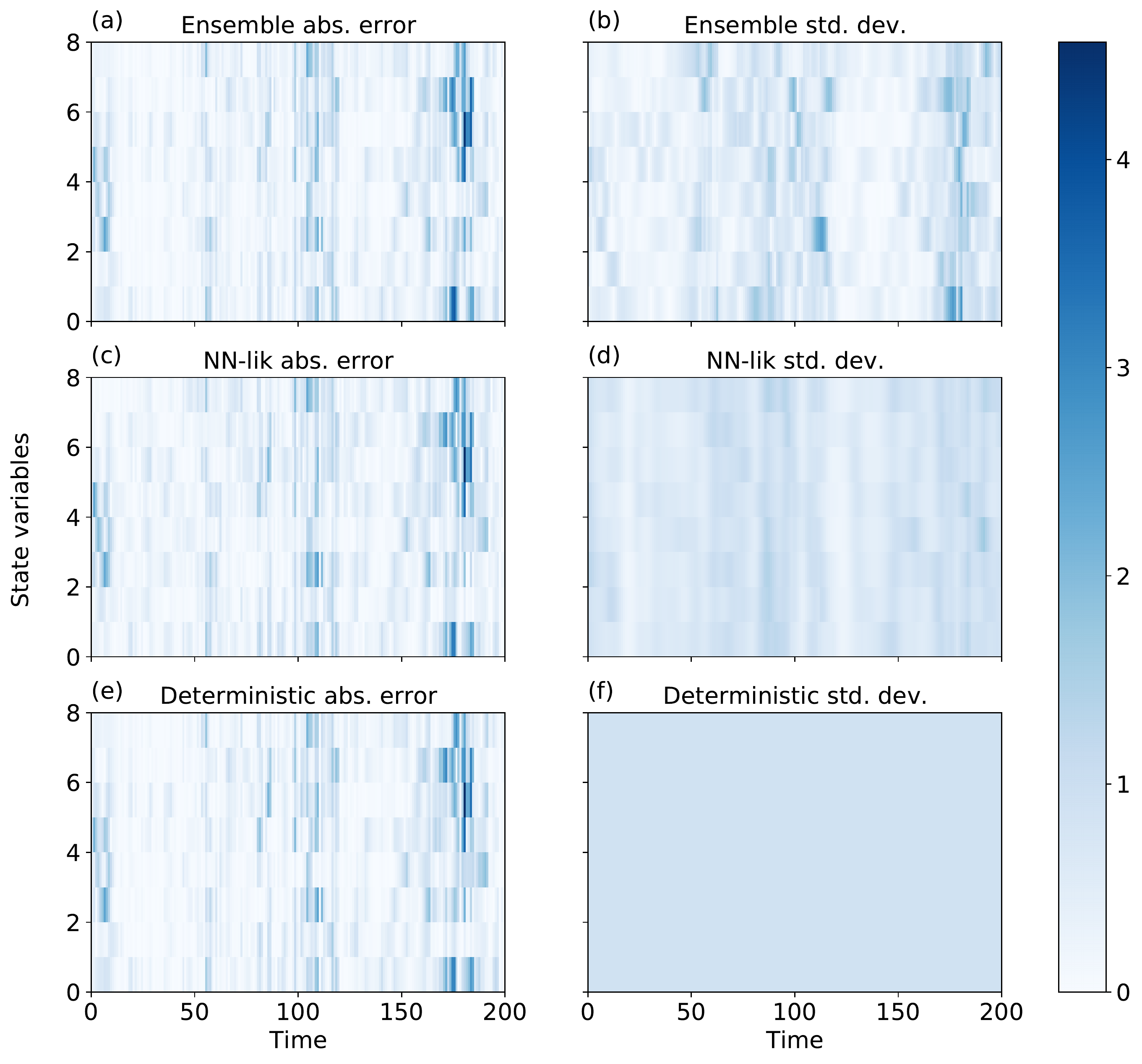} 
     \caption{Left panels (a,c,e): Time evolution of the absolute error for each state variable (y-axis) of the dynamic model at T80 lead time for the ensemble mean (a), the NN-lik approach (c), and the deterministic forecast (e). Right panels (b,d,f): the forecast error standard deviation as estimated from the ensemble (b), the NN-lik approach (d) and the deterministic forecast (f). }
    \label{fig:timeMXSE}
\end{figure}
 
Figure \ref{fig:timeMXSE} shows the spatio-temporal evolution of the absolute error ($ \left| {\v \epsilon} \right| $, left panels) and the estimated standard deviation of the forecast (right panels) for the ensemble, the NN-lik approach and the deterministic forecast. In the case of the deterministic forecast, its uncertainty is estimated as the standard deviation of the difference between the forecast and the analysis ($\v x^{d}_{l,k+l} - \v x^{a}_{0,k+l}$) over the training sample, and thus is independent of time (Figure \ref{fig:timeMXSE}f).
Absolute errors for the three experiments show a relatively large variability and are rather similar in terms of magnitude, spatio-temporal evolution and location of the extreme values. This confirms the strong dependence of forecast errors with the state of the system.

In terms of standard deviation prediction (Figs \ref{fig:timeMXSE}b, \ref{fig:timeMXSE}d and \ref{fig:timeMXSE}f), areas with maximum spread in the ensemble are in good agreement with the location of larger forecast errors. It is important to note that we do not expect a perfect match between these two quantities since the forecast error is one realization from a probability density function whose standard deviation is being approximated from the forecast ensemble. 
The ANN also provides an estimation of the forecast error standard deviation in good agreement with the evolution of the forecast error. As has been previously noted, the forecast standard deviation estimated from the network is smoother and shows less variability than the one estimated from the ensemble.  

These results are quite encouraging. They show that the NN-lik approach is able to estimate the forecast uncertainty as a function of the state of the system. Moreover, this estimation is achieved using a deterministic forecast as input whose computational cost is much cheaper than integrating an ensemble of forecasts. 

\begin{figure}[hbt!]
    \centering
     \includegraphics[width=\textwidth]{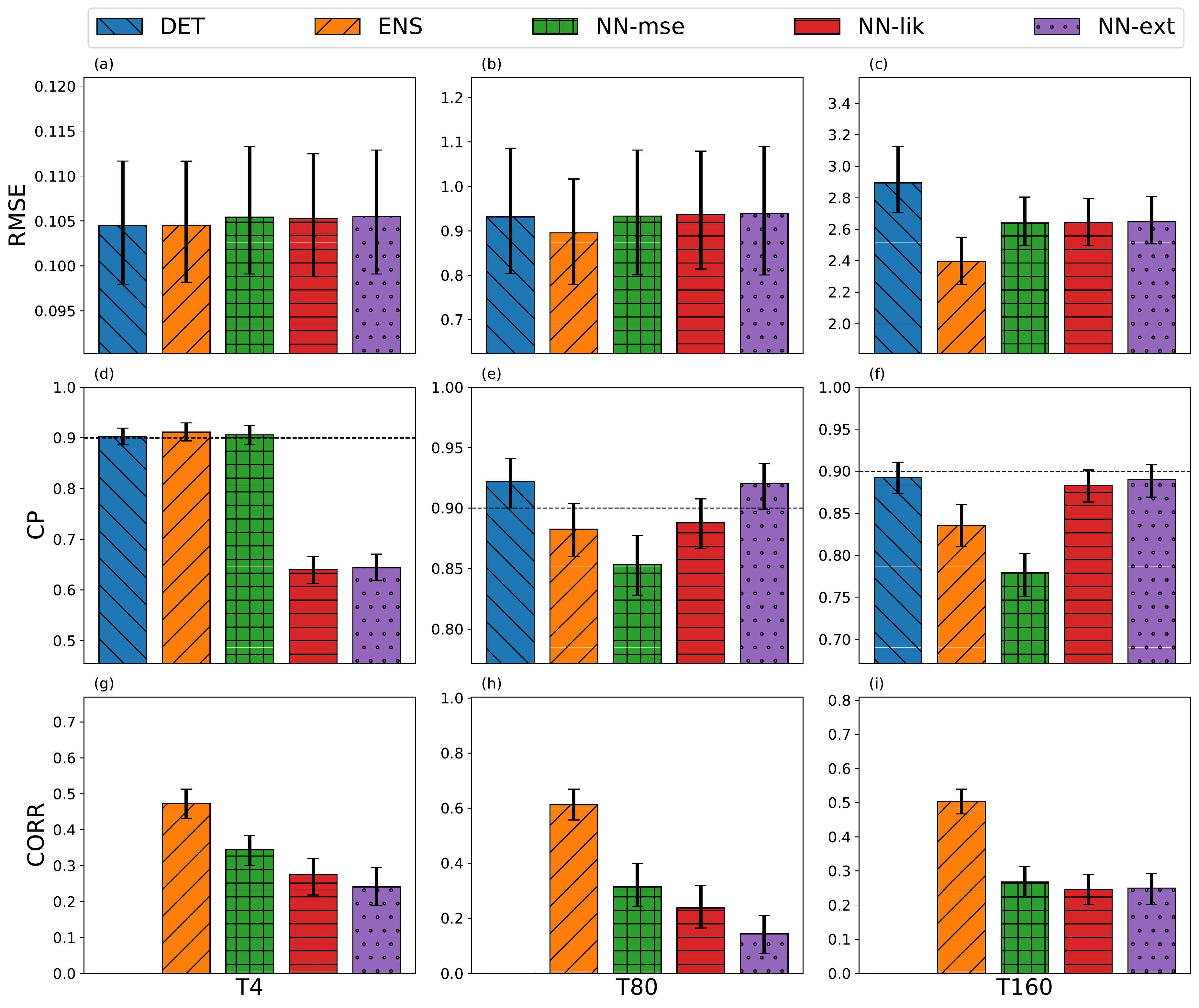}
     \caption{Scores corresponding to the deterministic forecast (blue bar), ensemble forecast (yellow bar), and the NN-mse (green bar), NN-lik (red bar) and NN-ext (purple bars) approaches in the perfect model scenario. The first row corresponds to the RMSE, the second to the 90\% coverage probability and the third to the standard deviation-absolute error correlation coefficient. The first, second and third columns corresponds to the T4, T80 and T160 lead times respectively. The error line on top of the bars indicates the 95\% confidence interval calculated using bootstrap technique. The horizontal line in the second row panels indicate the perfect value associated with the 90\% coverage probability.}
    \label{fig:estSE}
    
\end{figure}

So far we have qualitatively analyzed the skill of the ANN in quantifying the forecast uncertainty from a deterministic forecast at lead time T80. 
Figure \ref{fig:estSE} summarizes the results obtained for the different output lead times considered (T4, T80 and T160) and for the different scores described in Section \ref{SEC:METH}. Confidence intervals for the different scores are computed using a bootstrap approach in which 500 samples are obtained from the original sample using random selection with replacement. Forecast at valid times which are more than 20 time steps apart from each other are selected for the computation of the scores to increase the independence between different sample elements.  

Regarding the RMSE (first row of Fig. \ref{fig:estSE}),  all forecast systems show a similar performance for T4 and T80 with the ensemble mean performing slightly better at T80. All the ANNs have the same RMSE since the training of the network producing the corrected forecast is the same for all of them. 

For time T160 (Fig. \ref{fig:estSE}c) error growth is affected by nonlinear effects, the ensemble is able to adequately filter some of the unpredictable modes resulting in a lower RMSE with respect to the deterministic forecasts. One important result is that the ANNs outperform the deterministic forecast for this lead time. This suggests that ANNs are able to filter at least part of the unpredictable modes, although it is not as efficient as the ensemble mean in doing that. 

The second row in Fig. \ref{fig:estSE} shows the coverage probability (CP). A strong underestimation of the uncertainty is present in the estimation of the NN-ext and NN-lik approaches for the T4 forecast times. At lead time T4, the magnitude of the forecast error in PMS is highly correlated with the analysis error, namely the forecast error proxy used for training is underestimating the magnitude of the forecast error. This affects the NN-ext and NN-lik approaches which rely on this proxy for the estimation of the forecast uncertainty but not the MSE network which is trained against the uncertainty estimated by the ensemble.  
To verify this conjecture, we performed additional experiments training a network with the true state. As a result, an improvement in terms of RMSE and CP is found. In particular the sub-estimation of the forecast uncertainty at T4 is not present in these experiments. CP values went from 0.63 when training with analysis to 0.87 when training with ground truth, and RMSE values varies from 0.12 to 0.10 respectively.

At longer lead times (T80 and T160) the ensemble and the NN-mse network under-estimate the forecast standard deviation. The reason for this is not clear but it may be the result of a sub-optimal estimation of the analysis uncertainty in the ensemble-based data assimilation system. In any case, this behaviour shows a possible limitation of the NN-mse approach which can inherit biases in the quantification of the forecast uncertainty derived from a sub-optimal formulation of the ensemble used for the training of the network. 

The third row in Fig. \ref{fig:estSE} shows the spatio-temporal correlation between the estimated forecast standard deviation and the absolute forecast error. 
A low correlation between the error and the estimated forecast standard deviation indicates that the system is unable to capture the state-dependent nature of the forecast uncertainty. Figures \ref{fig:estSE}g, h and i show positive correlations for the ensemble forecast and the ANNs. As expected the time independent standard deviation of the error associated with the deterministic forecast is uncorrelated with the actual forecast error. The ensemble forecast outperforms the ANNs for all the considered lead times. This is a expected result since the ensemble forecasts is based on direct sampling from the probability density function of the state of the system, particularly in these perfect model experiments. 
The NN-mse network which is trained directly from an ensemble of forecasts performs slightly better than the indirectly trained networks at T4 and T80, but at T160 all ANNs shows similar performances. 
We consider other performance metrics such as the CRPS, however these are not shown since they do not provide additional information from that provided by the above discussed metrics. 

\subsection{IMS experiments}
\label{SEC:RIMS}
\begin{figure}[hbt!]
%    \centering
     \includegraphics[width=\textwidth]{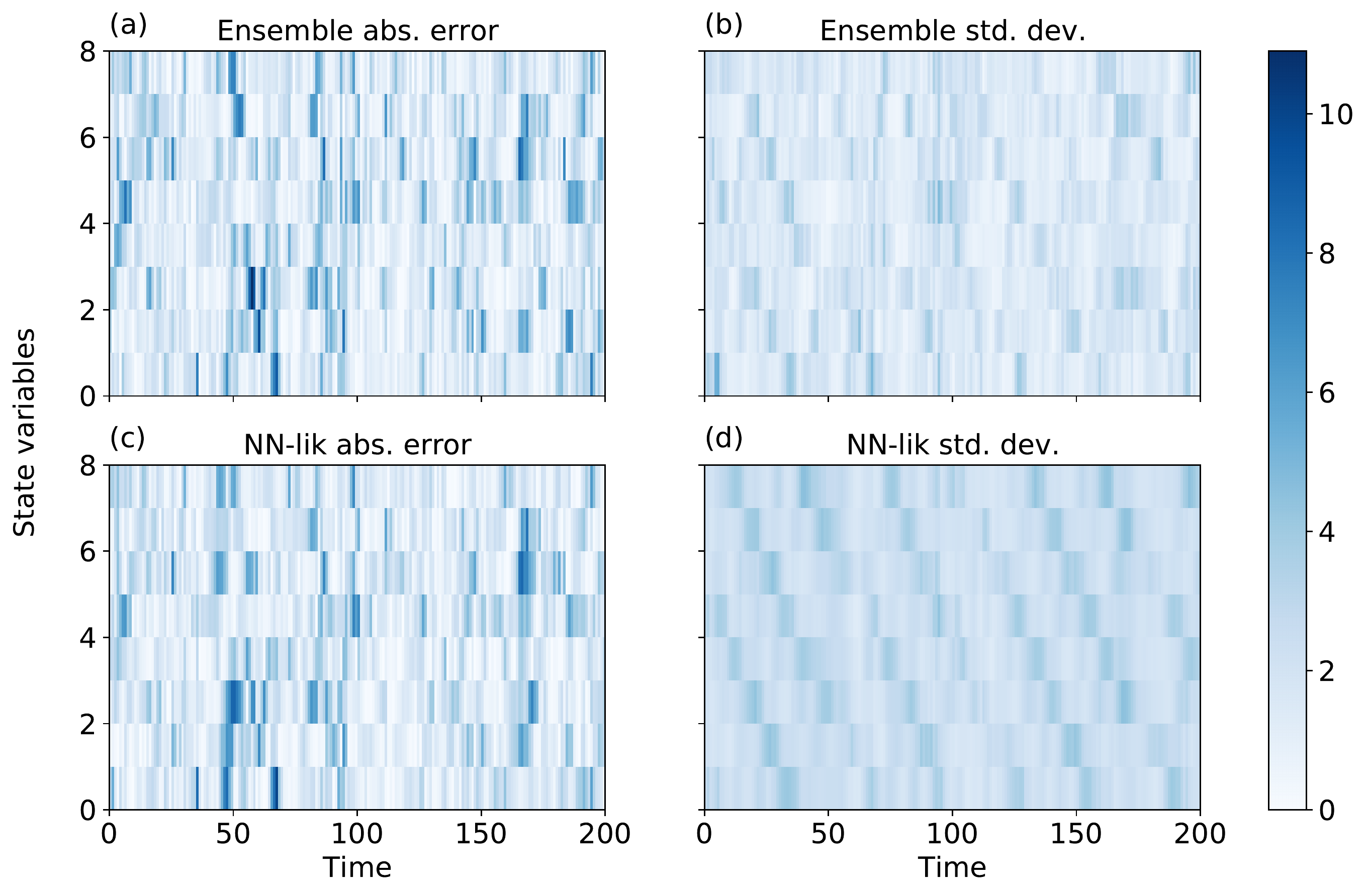} 
     \caption{Left panels (a,c):Time evolution of the absolute error for each state variable (y-axis) of the dynamic model at T80 lead time for the ensemble mean (a) and the NN-lik approach (c). Right panels (b,d): the forecast error standard deviation as estimated from the ensemble (b) and the NN-lik (d)}
    \label{fig:timeMXSSE}
\end{figure}

In the imperfect model scenario forecast errors are larger due to the contribution of model errors. Moreover, the  forecast error mean is usually non zero, i.e., there is a systematic error component. In this scenario the ANNs aim to \textit{"remove"} the systematic error component from state forecast and to capture state-dependent contributions of model errors to the forecast standard deviation.  

As shown,  the error grows more rapidly  in the IMS than in the PMS (see Figure \ref{fig:RMSE_LEAD}). In the IMS experiment, the T80 lead time is at the beginning of the nonlinear regime and T160 is near the error saturation meaning that T80 and T160 are representative of different model error regimes.  

Figure \ref{fig:timeMXSSE} shows the spatio-temporal evolution of the absolute error and its standard deviation estimated by the ensemble and the NN-lik approach. Although the error patterns are not as similar as those observed in PMS (Fig. \ref{fig:timeMXSE}), there is a strong similarity in shape and intensity (Fig. \ref{fig:timeMXSSE}a and \ref{fig:timeMXSSE}c).
The standard deviation panels, Fig. \ref{fig:timeMXSSE}b and \ref{fig:timeMXSSE}d,  show that both, the network and the ensemble, are able to represent the state-dependent forecast uncertainty as in the PMS experiment. 
The estimated forecast standard deviation is larger for the NN-lik approach than for the ensemble. This indicates a sub-estimation of the error standard deviation in the ensemble forecasting approach that may be partially attributed to the lack of a model error representation in the ensemble forecast experiment. 

\begin{figure}[hbt!]
        \includegraphics[width=\textwidth]{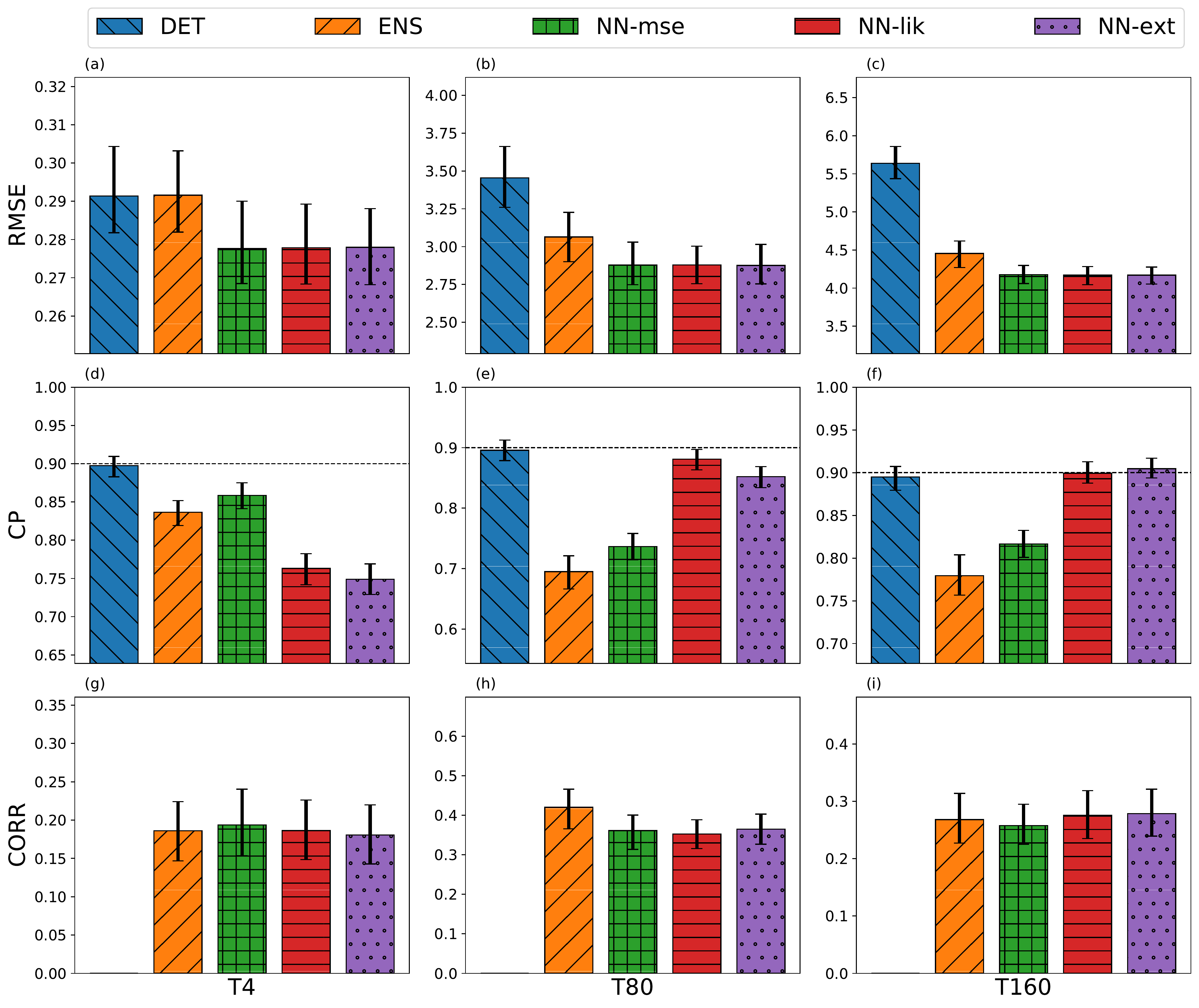}
    \caption{Same as Fig. 6 but for the imperfect model scenario. }
    \label{fig:estSSE}
\end{figure}

A summary of the performance of the different forecast systems in the IMS scenario is shown in Fig. \ref{fig:estSSE} for lead times T4, T80 and T160. In terms of the RMSE, a noticeable advantage in the accuracy of the state estimate is achieved by the ANNs at all lead times. The ANN corrected forecasts are even better than the ensemble mean forecast. This suggests that the contribution of model errors is significant and comparable or even larger to the error associated with the imperfect knowledge of the initial conditions. The ANNs are able to partially capture the state-dependent model errors while the deterministic forecast and the ensemble forecast are not. %(see Fig. \ref{fig:SCATML} and Fig. \ref{fig:SCATNN}).  

The CP (second row in Fig. \ref{fig:estSSE}) shows a clear advantage of ANNs over the deterministic and ensemble forecasts, particularly at T80 and T160. At T4 a similar behaviour is found as in the PMS with NN-ext and NN-lik performing worse than NN-mse. This is again a direct consequence of using the analysis as a learning target for the estimation of the forecast uncertainty. Once the forecast error has become larger (e.g. T80), this behavior is reversed: the networks that learn the uncertainty from the data perform better than the ensemble in terms of CPs shown  in Fig. \ref{fig:timeMXSSE}. 
It is interesting to note that the NN-mse learning technique, which learns directly from the ensemble spread performs better than the ensemble itself in terms of CP. This is partially because, the network improves the forecasted state by removing the state-dependent systematic forecast error component. Thus, the estimated forecast standard deviation is more consistent with the actual forecast errors than in the case of the ensemble forecast. This suggests that even an ensemble with a sub-standard representation of model errors can be a reasonable target for the training of an ANN, although it is not clear whether this particular result generalizes to more complex model errors as the ones present in state of the art numerical weather prediction or climate models \citep{bonavitaetal2020}.  
The ensemble shows the worst performance in terms of CP for T80 and T160, even worse than the deterministic forecast. This is because the ensemble mean is affected by model errors which are not accounted for in the formulation of the ensemble leading to an under-estimation of the forecast standard deviation.

The correlation between the absolute error and the forecast standard deviation (third row in Fig. \ref{fig:estSSE}) shows that the ANNs in the IMS perform as well as the ensemble forecast. 

\begin{figure}[hbt!]
%    \centering
        \includegraphics[width=\textwidth]{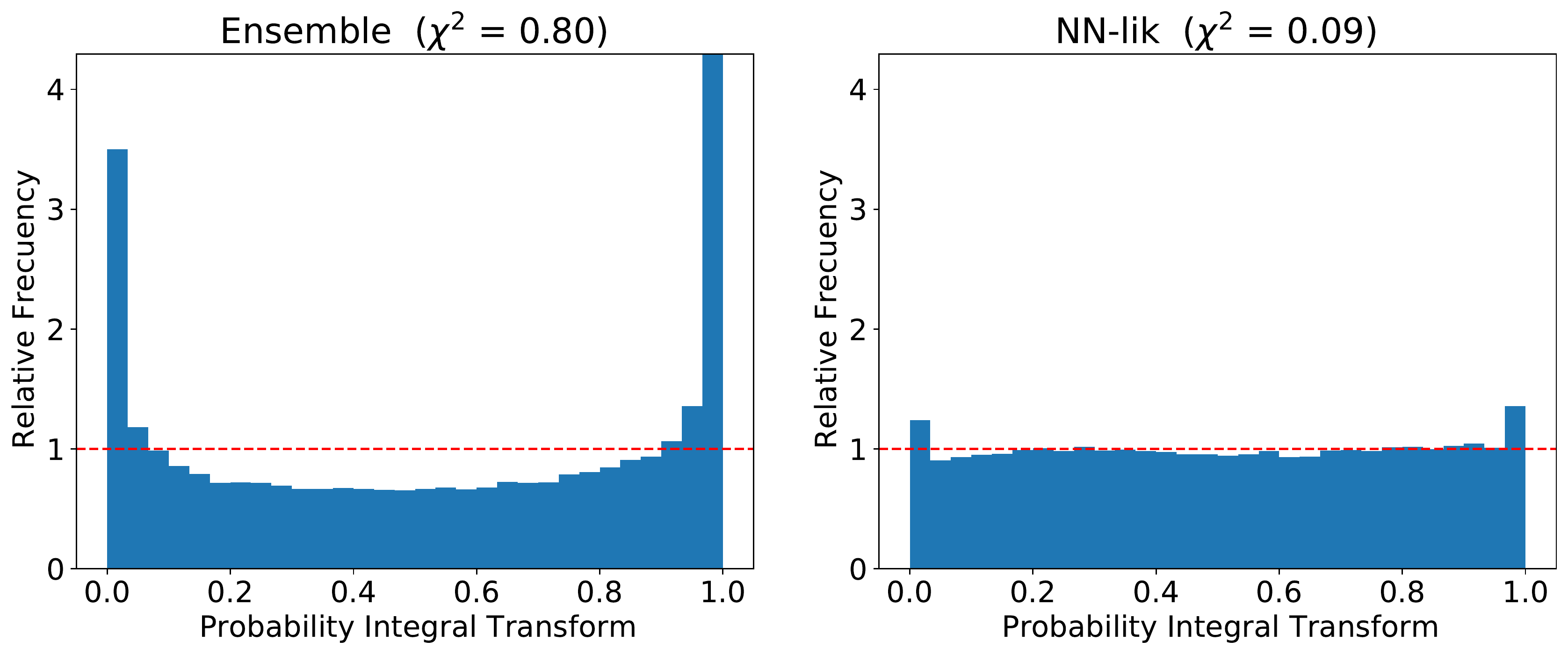}
        \caption{Probability integral transform histogram for the ensemble forecast (left panel) and for the NN-lik forecast. Both cases corresponds to the T80 forecast lead time.The corresponding $\chi^2$ flatness parameter is indicated on top of each panel.}
    \label{fig:PITH}
\end{figure}

Figure \ref{fig:PITH} shows the PIT histograms for forecasts at T80 generated by the ensemble and NN-lik. The other NN training approaches perform similarly and are not included in the figure. The ensemble forecast exhibits a U-shaped PIT histogram, indicating an underestimation of the forecast uncertainty. This lack of reliability in the ensemble is associated to a sub-optimal representation of model errors that introduce systematic biases in the ensemble mean and contributes to the stochastic component of the forecast errors. Both aspects are partially taken into account by the network explaining the improvement of the forecast reliability (note that value of $\chi^2$ is an order of magnitude smaller for the networks than for the ensemble).

\subsection{Estimation performance analysis}

\begin{figure}[hbt!]
%    \centering
    \includegraphics[width=\textwidth]{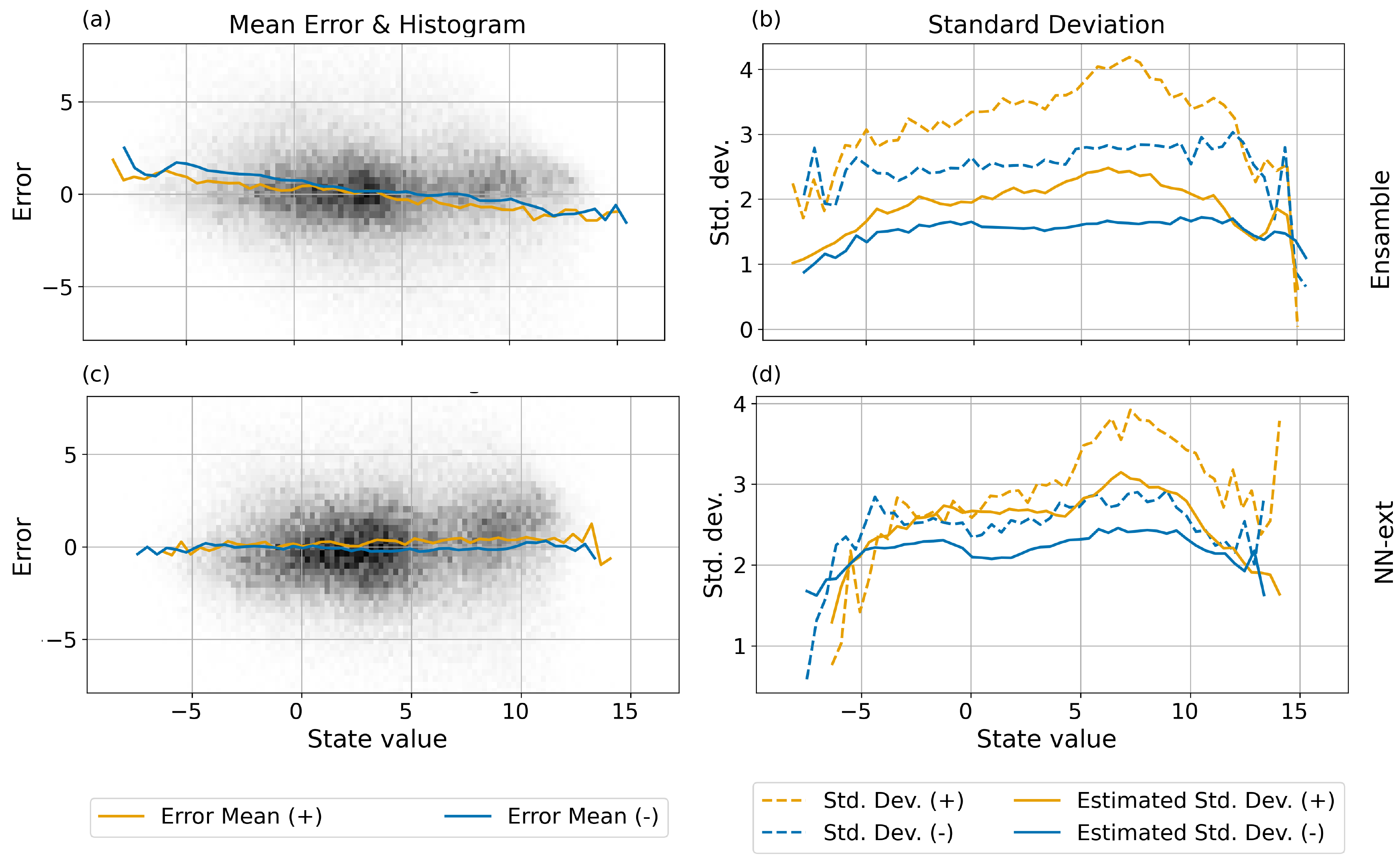}
    \caption{Panels (a) and (c) show the 2D density of the forecast error and forecasted state values (grey shades). The yellow and blue solid lines indicates the mean forecast error as a function of the forecasted state value for the positive trend and negative trend sub-samples respectively (see the text for details). Panels (b) and (d) show the standard deviation of the error as function of the forecasted state variable for the positive trend and negative trend sub-samples (dashed yellow line and dashed blue line respectively). The solid yellow and blue lines represents the estimated standard deviation for the positive trend and negative trends sub-samples respectively. Panels (a) and (b) corresponds to the ensemble forecast and (c) and (d) to the NN-ext forecast. All panels corresponds to the imperfect model scenario.}
    \label{fig:SCAT}
\end{figure}

To gain more insight into how forecast errors depend on the state variables and to better understand to what extent these dependencies are represented by the ANNs and the ensemble, we examine some aspects of the relationship between state variables and the forecast error. We focus on the IMS experiments discussed in the previous section. 
Left panels in Fig. \ref{fig:SCAT} show  2D histograms of the forecast error distribution at each particular variable (i.e. $\epsilon_{(i)}$) as a function of the value of the forecasted state variable for the the ensemble mean forecast ($\bar { \v x}_{(i)}^e$) and the NN-ext ($\tilde {\v x}_{(i)}$). Since for this simple model all state variables (subscript $_{(i)}$) share the same error statistics, results from different state variables are pooled together to construct a single sample. We investigate the relationship between the error and the forecast state at the corresponding variable under two different regimes: One in which the trend at the corresponding state variable is positive (i.e. $\frac{dx_{(i)}}{dt} > 0$) and one in which the trend at the corresponding state variable is negative (i.e. $\frac{dx_{(i)}}{dt} < 0$). This distinction is motivated by the work of \cite{pulido16}, that shows, in a similar experimental setting, that the magnitude of model errors has a non-Markovian behavior which depends on the time derivative of the state variables.

The solid lines over the 2D histogram in Figs. \ref{fig:SCAT}a and \ref{fig:SCAT}c, represent the mean error as a function of the state value for the positive trend sub-sample (yellow line) and the negative trend sub-sample (blue line). To construct these curves, the range of  state values is divided into 50 bins, and then the mean of the forecast errors corresponding to each bin is computed. This procedure is conducted independently for the positive trend and negative trend sub-samples.  

Figs. \ref{fig:SCAT}b and \ref{fig:SCAT}d show the error standard deviation as a function of the forecasted state value (dashed lines). These curves are obtained independently for the positive trend sub-sample (yellow dashed line) and for the negative trend sub-sample (blue dashed line) following a similar procedure as for the mean error curves.
The different estimations of the error standard deviation are shown as solid lines in Figs. \ref{fig:SCAT}b and \ref{fig:SCAT}d. In this case we compute the average of the estimated error standard deviation at each of the bins covering the forecasted state variable range. This procedure is also applied independently to the positive trend (yellow solid line) and negative trend (blue solid line) independently.

The ensemble mean error strongly depends on the state variable (slope in the mean error), although the globally averaged error is close to 0 (Fig. \ref{fig:SCAT}a). This slope is likely associated with the following mechanism: when a high value of the state variable is forecasted, it is more likely to have a negative error than a positive one (related to a subestimation of the value). The opposite is true for low values of $x$. This produces an asymmetry in the shape of the error probability distribution that becomes more evident for extreme state values. Instead, the NN-ext mean error (Fig. \ref{fig:SCAT}c) is almost independent of the forecasted state value. This behaviour comes at the expense of less frequent in the ANN corrected forecasts.  Note that the histograms in Fig. \ref{fig:SCAT}c  span a smaller range in state values than in Fig. \ref{fig:SCAT}a. 

A clear gap between the positive and negative trend sub-samples is found in Fig. \ref{fig:SCAT}b for the ensemble standard deviation which also depends on the forecasted value. This result is consistent with the findings of \cite{pulido16}.
The ensemble spread (Fig. \ref{fig:SCAT}b) underestimates the standard deviation of the forecast error but adequately represents its dependence with the forecasted value. 

Figure \ref{fig:SCAT}d shows that the NN-ext provides a more reliable estimation of the forecast standard deviation compared with the ensemble (solid and dashed lines are closer) and also capture the relationship between the forecast standard deviation and the forecasted state value. 
These results suggests that the dependence of the uncertainty on the state of the system is complex and non-linear and that NN-based uncertainty quantification is able to capture these complexity in a similar way as an ensemble of forecasts.

\section{Real data cases analysis}
\label{SEC:RDC}
To investigate the performance of NN-based uncertainty quantification in more realistic scenarios, we train an NN using an uncertainty-aware loss function to estimate the error variance of deterministic forecasts produced by a state-of-the-art global numerical weather prediction system.
In this section we will show how results similar to those analyzed in the previous sections for the Lorenz96 model are also verified in at least one simple real case

\subsection{Dataset}
In this experiment we used the deterministic sea level pressure forecasts from the National Centers for Environmental Prediction (NCEP) reforecast version 2 \citep{guanetal2022} available from January $1^{st}$ 2000 to December $31^{st}$ 2019.Reforecasts are retrospective forecasts performed with the stable version of the numerical weather prediction model which is close to the most recent version used in operations. This procedure eliminates changes in the systematic error components associated with evolving model versions. This dataset allows for a better characterization of systematic model errors and their dependence with the state of the system, making them particularly appealing for the implementation of approaches like the ones described in this paper.

This datasets consists of a global 5-member ensemble (one control forecast and 4 perturbed forecasts). The forecasts are initialized once a day at 00 UTC, and are stored with a spatial resolution of 0.25 degrees and a temporal resolution of 3 hours. 
We perform the experiments on a domain located over the South Atlantic (55$^\circ$ W to 5$^\circ$ W and 60$^\circ$ S to 40$^\circ$ S). This area is located within the mid-latitudes westerly flow and is quite active in terms of baroclinic waves.Consequently a significant variability in the forecast uncertainty is expected in this area. To reduce the dimensions of the problem, we upscaled the resolution of the forecast to 2$^\circ$ resulting in a domain of $10 \times 10$ grid points.
In this experiment we aim to obtain two NNs providing an improved forecast and its associated uncertainty at a lead time of 72 hours. This lead time is in the linear phase of the error growth.  The input to the networks is a sequence of deterministic forecasts (i.e. the control forecast) at 0, 36 and 72hs lead time. The initial conditions of the forecast are used as the training target.

The entire reforecast dataset consists of 7302 forecasts. We used 5096 for training, 1104 for validation and 1102 for testing.
To reduce the impact of seasonal variations on pressure variability and on the uncertainty associated to its forecast, we standardized pressure values with respect to their climatological mean and standard deviation. To filter the seasonal cycle we compute the mean and standard deviation over a time window of 30 days centered around each day of the year.

\subsection{Neural Network}
In this experiment, we use the same architecture as in the rest of the paper: two fully connected network with two hidden layers, one for the estimation of the corrected forecast and the other for the estimation of the standard deviation of the forecast error. The size of the input and output are determined by the grid size and consist of $3\times10\times10$ neurons  and $10\times10$ respectively for both  networks.

We explore the sensitivity of the performance of the network to the size of the hidden layers in the range of 100 to 2000 neurons finding little improvements beyond 500 neurons.

For the training of the networks the same methodology was followed as explained in Section \ref{SEC:METH}.

\subsection{Results}

\begin{figure}[hbt!]
%    \centering
        \includegraphics[width=\textwidth]{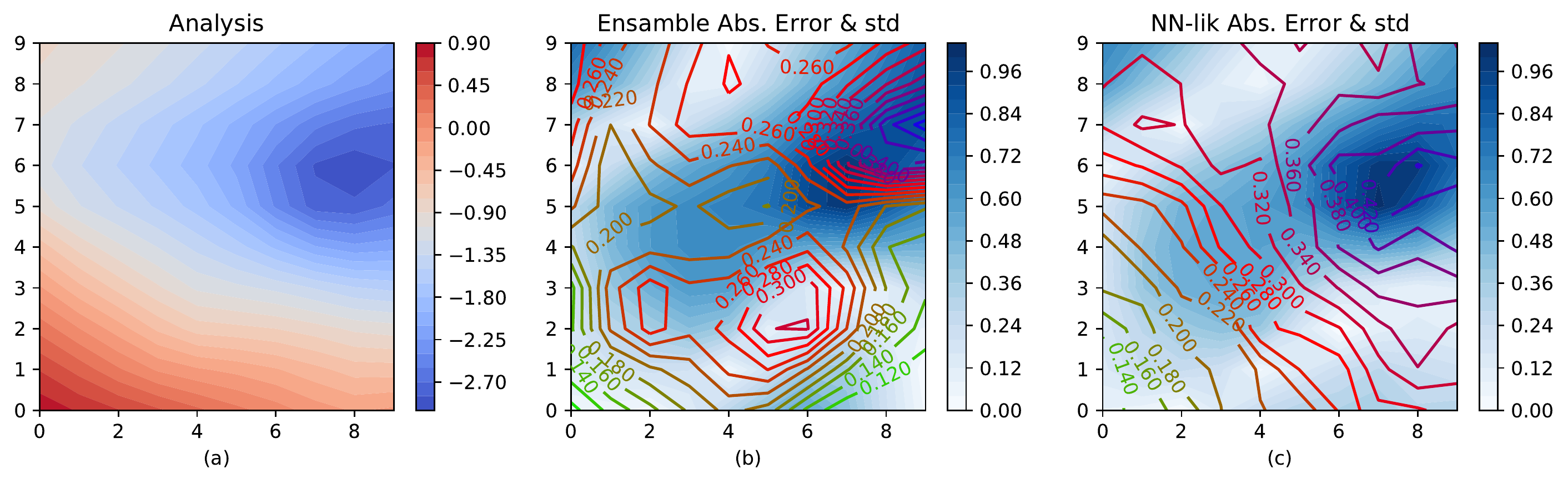}
        \caption{Panel (a) shows the analyzed spatial distribution of the sea level pressure corresponding to the 20 of June of 2019. Panels (b) and (c) show the 72 hs lead time forecast error (shaded) and the estimated error standard deviation (contours) as provided by the ensemble (b) and the neural network (c)}
    \label{fig:REALAES}
\end{figure}

Figure \ref{fig:REALAES}a shows the analyzed state corresponding to June $20^{th}$ of 2019 corresponding to an intense extratropical cyclone. Both the ensemble and the network place the maximum standard deviation of the error at and to the east of the cyclone. The spatial distribution of the uncertainty estimated by the NN is smoother than the one obtained from the ensemble. For the NN there seems to be a dependence of the uncertainty with the surface pressure value, with higher uncertainties associated with low pressure systems. This is expected since low pressure systems exhibit larger deepening rates and are also associated to small scale phenomena (such as deep moist convection) whose effect can not be accurately represented in the state-of-the-art global numerical models, resulting in a larger contribution of model errors to the total forecast error. 

\begin{figure}[hbt!]
%    \centering
\includegraphics[width=\textwidth]{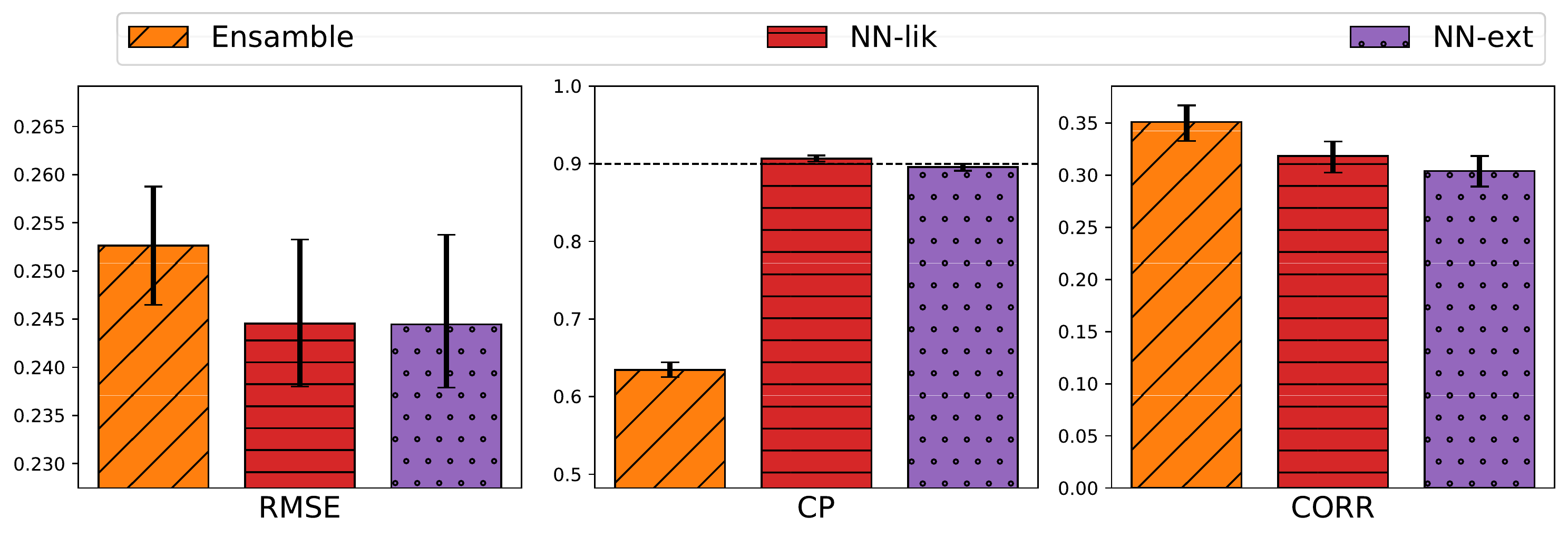}
        \caption{Scores corresponding to ensemble forecast (yellow bar), and the NN-lik (red bar) and NN-ext (purple bars) for the real data case for root mean squared error (right panel), the 90\% coverage probability (center panel) and the standard deviation-absolute error correlation coefficient (left panel). The error bar indicates the 95\% confidence interval calculated using the bootstrap technique. The horizontal line in the second panel indicates the perfect value associated with the 90\% coverage probability.}
    \label{fig:estREAL}
\end{figure}

Figure \ref{fig:estREAL} shows the RMSE, coverage probability and correlation between the forecasted error standard deviation and the absolute error for the ensemble and for the NN-lik and NN-ext. These metrics show a similar behaviour as in the synthetic experiments with the imperfect model at T80. Unlike the idealized experiments, in this case we do not have access to the true state, so we use the analysis as an estimation of the state of the system to evaluate the performance of the networks.

The networks show an improvement in the root mean square error (RMSE), a better reliability as measured by the coverage probability (CP). The NNs also shows a much flatter PIT histogram ($\chi^2$ values of 0.15 for both the NN-lik and NN-ext versus 1.03 for the ensemble forecast). However, both NNs are outperformed by the ensemble in terms of the correlation. It is also observed that the NN-lik network gives slightly better results than NN-ext as in the synthetic cases.

\section{Conclusion and perspectives}
\label{SEC:CONC}

In this work we evaluate different strategies for the joint estimation of state-dependent systematic forecast error and the forecast uncertainty using machine learning. We compare the performance of artificial neural networks (ANNs) that estimate the forecast uncertainty and that are directly trained from an ensemble forecast or indirectly trained using  loss function formulations that implicitly take into account the uncertainty. We implement two different approaches for the implicit training, one based on a local uncertainty estimation loss function (referred to as NN-ext) and the other based on a Gaussian likelihood loss function (referred to as NN-lik). The input to the ANNs is a set of deterministic forecasts at different lead times. We evaluate the performance of the ANN from experiments using a chaotic dynamical model, and compare them with a deterministic forecast and an ensemble forecast in perfect model (PMS) and imperfect model scenarios (IMS)  at different forecast lead times. 

Overall the results obtained in this work in a simple proof-of-concept experimental setting are quite encouraging about the possibility of increasing the accuracy of a deterministic forecast and providing a reliable estimation of its uncertainty at a much lower computational cost than that required by an ensemble forecast. 
In particular, we show that the ANNs are able to filter part of the uncertainty associated to unpredictable modes, particularly in the nonlinear error growth regime. In this aspect, the ANNs are able to emulate the filtering effect of the ensemble mean, although with less accuracy. The ANNs are also able to accurately represent the state-dependent forecast error standard deviation resulting from initial conditions errors and their amplification due to the chaotic dynamics of the system. A systematic sub-estimation of the forecast standard deviation is found at short lead times in the ANNs due to our training strategy in which the analysis is used as a proxy of the true state of the system. 

When the model is imperfect, the ANNs are capable of estimating the state-dependent systematic error component associated with the effect of model errors.  
This result is particularly encouraging since a proper representation of model errors and its impact upon the forecast remains a significant scientific challenge. 
The network directly trained with an ensemble of forecasts performs better in the perfect model scenario. However, the indirect training approaches (NN-ext and NN-lik) produce better results in the imperfect model scenario. This is mainly because in the direct training, the network inherits the uncertainty estimation achieved by the ensemble and since model errors are usually sub-represented, this may lead to worse results in comparison to a network which is indirectly trained and able to learn the effect of model errors from the data.

Additionally, in this work, we present proof-of-concept experiments using a state-of-the-art global numerical weather prediction system. Results show a similar performance as the one obtained in the simple model experiments suggesting that the main aspects of the methodology can be extended to more complex systems with some  cautions. One important aspect related to the extension of these methodologies to state-of-the-art systems is the increase in the dimension of the inputs and the outputs. In our experiments we use simple network architectures that do not scale well with the size of the input-output of the network and of the hidden layers. In order to conduct an NN-based uncertainty quantification with higher resolution models over larger domains other architectures should be considered such as deep convolutional networks (e.g. \cite{gronquist2021,chapmanetal2022}). Another possibility to deal with the course of dimensionality is to work on local domains as has been extensively done in the field of data assimilation \citep{carrassi2018}. 
More research is also required to investigate how analog methods \citep{platzer2021,alessandrinietal2019} and other advanced machine learning methodologies are potentially able to achieve better results in more realistic scenarios, and which strategies can be used to reduce the dimensionality of the uncertainty estimation problem.

One direct application of our results is the development of hybrid data assimilation systems in which ANNs can produce an estimation of the uncertainty of the forecast at a lower computational cost than that required by ensemble forecasts. This is in fact a more challenging scenario since an estimation of the full error covariance matrix is required in this context which implies a significant increase in the dimensionality of the estimation problem. The development of a hybrid machine-learning based data assimilation approach based on the results presented in this paper is part of a work in progress.

\section{acknowledgments}

We thank the ECOS-Sud Program for its financial support through the project A17A08. This research has also been supported by the National Agency for the Promotion of Science and Technology of Argentina (grant no. PICT-2233-2017, PICT-SERIEA-01168, PICT-2021-CAT-I-130), the University of Buenos Aires (grant no. UBACyT-20020170100504). Special thanks to the National Meteorological Service of Argentina for their support and trust in this work.

\bibliography{references}
\end{document}